\documentclass{article} 
\usepackage{iclr2027_conference,times}

\usepackage{amsmath,amsfonts,bm}

\def\eqref#1{equation~\ref{#1}}

\def\1{\bm{1}}

\DeclareMathAlphabet{\mathsfit}{\encodingdefault}{\sfdefault}{m}{sl}
\SetMathAlphabet{\mathsfit}{bold}{\encodingdefault}{\sfdefault}{bx}{n}

\usepackage{hyperref}
\usepackage{url}
\usepackage{graphicx,enumitem,wrapfig,hyperref,cleveref,booktabs,makecell,colortbl,xcolor,multicol,multirow,adjustbox,array,pifont,tcolorbox}
\usepackage{titletoc}

\titlecontents{section}
  [2em]
  {\addvspace{0.5em}\normalfont\normalsize\scshape}
  {\contentslabel{2em}\color[HTML]{397CB8}}
  {\color[HTML]{397CB8}}
  {\color{black}\titlerule*[0.5pc]{.}\contentspage}
\titlecontents{subsection}
  [4.5em]
  {\normalfont\small\scshape}
  {\contentslabel{3em}\color[HTML]{397CB8}}
  {\color[HTML]{397CB8}}
  {\color{black}\titlerule*[0.5pc]{.}\contentspage}

\tcbuselibrary{skins,breakable}
\usepackage[normalem]{ulem}
\usepackage[dvipsnames]{xcolor}
\definecolor{cYes}{HTML}{43A047}       
\definecolor{cNo}{HTML}{D32F2F}        
\newcommand{\cmark}{\textcolor{cYes}{\ding{51}}}     
\newcommand{\xmark}{\textcolor{cNo}{\ding{55}}}%
\definecolor{DeltaGreen}{RGB}{170, 225, 185}
\definecolor{DeltaRed}{RGB}{245, 180, 180}
\definecolor{promptbg}{HTML}{F7F9FC}     
\definecolor{promptframe}{HTML}{D0D7E2}  
\definecolor{prompttitle}{HTML}{4A6FB3}  
\newtcolorbox{promptbox}[2][]{
  enhanced,
  breakable,
  width=\columnwidth,  
  colback=promptbg,
  colframe=promptframe,
  coltitle=white,
  colbacktitle=prompttitle,
  fonttitle=\bfseries,
  title={#2},
  boxrule=0.8pt,
  arc=3mm,
  outer arc=3mm,
  left=2mm,
  right=2mm,
  top=2mm,
  bottom=2mm,
  #1
}

\title{IntReal: A Real-World Dataset and Benchmark for Multimodal Interleaved Generation}

\iclrfinalcopy

\author{\makebox[\textwidth][c]{Shaohan Yu$^{1,2}$ Jie Qin$^2$ Wende Tan$^3$ Haiwen Diao$^4$ Limeng Qiao$^2$ Yueming Lyu$^1$ Chenyang Si$^{1\dagger}$}\\
\makebox[\textwidth][c]{$^1$PRLab, Nanjing University\quad $^2$Meituan}\\
\makebox[\textwidth][c]{$^3$Imperial College London\quad $^4$S-Lab, Nanyang Technological University}\\
\makebox[\textwidth][c]{\texttt{shaohan.yu@smail.nju.edu.cn} \quad \texttt{chenyang.si@nju.edu.cn}} \\
\makebox[\textwidth][c]{\hypersetup{hidelinks}\color[HTML]{397CB8}\url{https://yushaohan.github.io/IntReal/}}
}

\begin{document}

\maketitle
\begin{figure*}[h!]
\vspace{-22pt}
\centering
   \includegraphics[width=\linewidth]{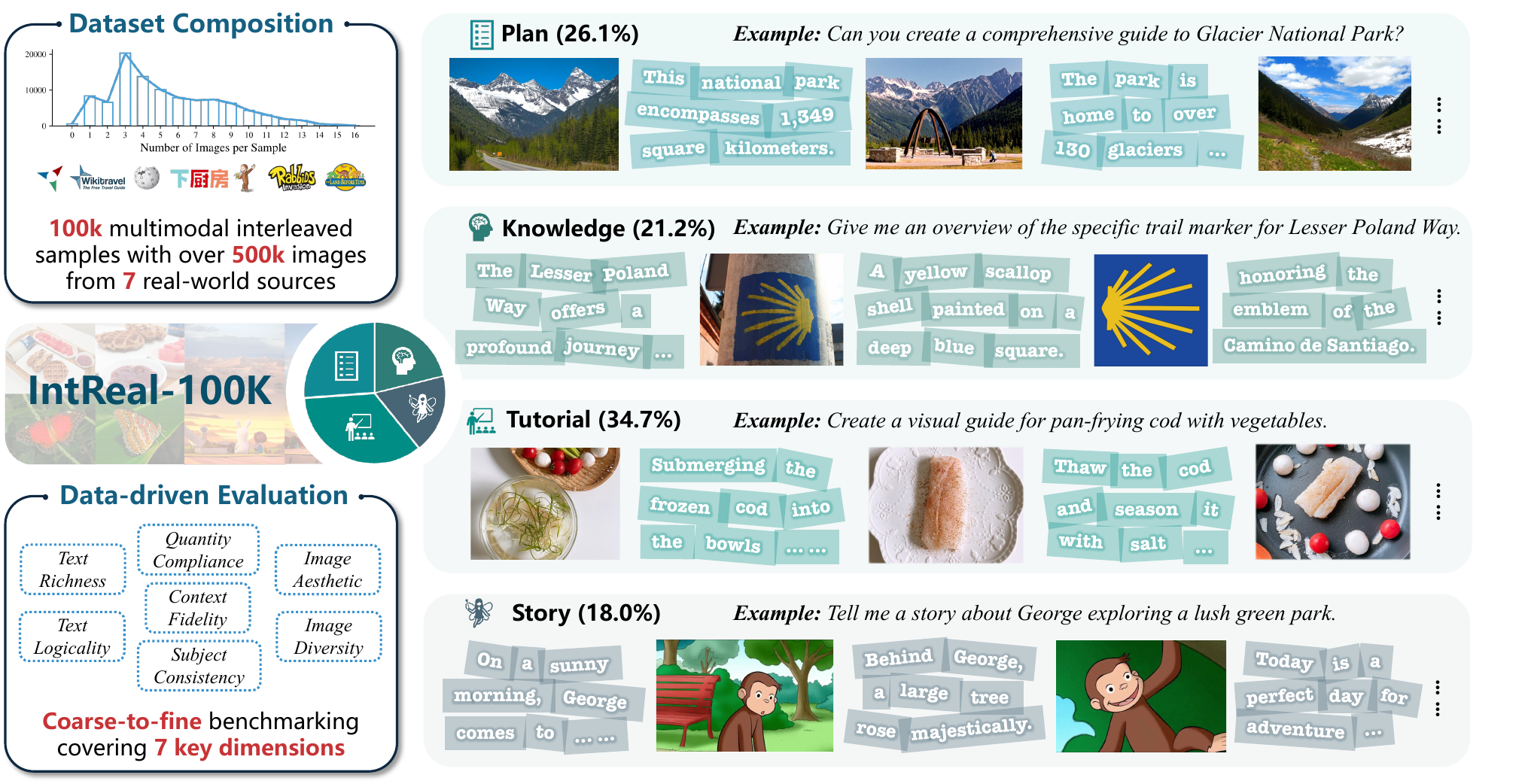}
    \caption{\textbf{Overview of IntReal.} IntReal-100K contains 100,000 multimodal interleaved samples with 538,452 images, curated from 7 real-world sources and organized into 4 categories. The IntReal Evaluation Suite performs coarse-to-fine diagnosis via 7 fine-grained dimensions.}
   \label{fig:teaser}
\end{figure*}
\begin{abstract}
Humans routinely consume multimodal content, where textual and visual information is naturally interleaved. Although Unified Multimodal Models (UMMs) can jointly generate text and images, multimodal interleaved generation remains largely underexplored. Progress in this area is hindered by two key limitations: (1) the lack of well-defined interleaved generation tasks and corresponding training data, and (2) the absence of fine-grained evaluation protocols that can reliably diagnose model performance across different capability dimensions. To this end, we present \textbf{IntReal}, comprising two complementary components: \textbf{IntReal-100K}, a real-world dataset of multimodal interleaved samples spanning four major categories, and the \textbf{IntReal Evaluation Suite}, which evaluates model performance across seven complementary dimensions using specialized expert models, enabling coarse-to-fine diagnosis. Extensive experiments validate our evaluation suite through low inter-metric redundancy (max pairwise absolute Spearman's $\rho \leq 0.42$), high agreement with human judgments (83.3\% accuracy), and strong consistency across different expert evaluators (Spearman's $\rho > 0.9$), supporting the complementarity and reliability of our metrics. By benchmarking four families of UMMs, we show that IntReal Evaluation Suite reveals distinct strengths and weaknesses across models. Post-training experiments demonstrate the utility of IntReal-100K for improving interleaved generation and further show that \textbf{training unimodal generation can also benefit interleaved generation}. These findings suggest that interleaved generation can benefit from modality-specific optimization, rather than requiring joint text-image training at every stage.
\end{abstract}

\section{Introduction}
\label{sec:intro}
\begin{table*}[t]
    \centering
    \begin{minipage}[t]{0.47\textwidth}
        \centering
        \large
        \setlength{\tabcolsep}{3pt}
        \caption{\textbf{Comparison with open-source interleaved image-text datasets.}
        IntReal combines understanding- and generation-oriented filtering with a taxonomy of four real-world scenarios and curated web and video sources.}
        \begin{adjustbox}{max width=\linewidth}
        \begin{tabular}[t]{lcccc}
            \toprule
            \multirow{2}{*}{\textbf{Dataset}}
            & \multicolumn{2}{c}{\textbf{Data Filtering}}
            & \multirow{2}{*}{\textbf{Taxonomy}}
            & \multirow{2}{*}{\makecell{\textbf{Data}\\\textbf{Source}}} \\
            \cmidrule(lr){2-3}
            & \textbf{For Und.} & \textbf{For Gen.} & & \\
            \midrule
            MMC4 & \cmark & \xmark & \xmark & Web \\
            OBELICS & \cmark & \xmark & \xmark & Web \\
            OmniCorpus & \cmark & \xmark & \xmark & Web \\
            CoMM & \cmark & \cmark & \xmark & Web \& Vid. \\
            \textbf{IntReal} & \cmark & \cmark & \cmark & Web \& Vid. \\
            \bottomrule
        \end{tabular}
        \end{adjustbox}
        \label{tab:dataset-comparison}
    \end{minipage}
    \hfill
    \begin{minipage}[t]{0.49\textwidth}
        \centering
        \large
        \setlength{\tabcolsep}{2pt}
        \caption{\textbf{Comparison with open-source multimodal interleaved generation benchmarks.} IntReal Evaluation Suite employs VLMs and expert models to provide coarse-to-fine evaluation.}
        \begin{adjustbox}{max width=\linewidth}
        \begin{tabular}[t]{lcccc}
            \toprule
            \multirow{2}{*}{\textbf{Benchmark}}
            & \multicolumn{2}{c}{\textbf{Evaluation Models}}
            & \multicolumn{2}{c}{\textbf{Evaluation Granularity}} \\
            \cmidrule(lr){2-3}
            \cmidrule(lr){4-5}
            & \textbf{VLM}
            & \makecell{\textbf{Expert}\\\textbf{Model}}
            & \textbf{Holistic}
            & \makecell{\textbf{Modality-}\\\textbf{Specific}} \\
            \midrule
            OpenING & \cmark & \xmark & \cmark & \xmark \\
            InterleavedEval & \cmark & \xmark & \cmark & \xmark \\
            ISG-Bench & \cmark & \xmark & \cmark & \cmark \\
            UEval & \cmark & \xmark & \xmark & \cmark \\
            \textbf{IntReal} & \cmark & \cmark & \cmark & \cmark \\
            \bottomrule
        \end{tabular}
        \end{adjustbox}
        \label{tab:benchmark-comparison}
    \end{minipage}
\end{table*}

Flip through a book or browse a webpage, and text and images are often tightly interleaved. Encyclopedic articles interleave text and images to communicate knowledge, storybooks integrate narratives with vivid illustrations, tutorials pair step-by-step instructions with corresponding visual demonstrations, and travel guides present itineraries alongside maps and photographs. Despite the emergence of Unified Multimodal Models (UMMs) capable of generating both text and images, multimodal interleaved generation remains largely treated as a by-product of model unification rather than a fundamental generative task in its own right~\citep{bagel,emu3p5,sensenova-u1}.

A major reason for this underexploration is the lack of datasets and benchmarks specifically designed for this task. On the data side, as shown in~\cref{tab:dataset-comparison}, large-scale interleaved datasets such as MMC4~\citep{mmc4}, OBELICS~\citep{obelics} and OmniCorpus~\citep{omnicorpus} are primarily collected for multimodal understanding rather than generation, and therefore do not provide generation-oriented quality control. CoMM~\citep{comm} aggregates interleaved data sources, but does not organize data collection around a taxonomy of real-world scenarios, leaving the task space of multimodal interleaved generation insufficiently characterized. This field thus lacks a high-quality interleaved generation dataset constructed under a principled taxonomy.  On the evaluation side, as shown in~\cref{tab:benchmark-comparison}, existing benchmarks often rely on vision-language models (VLMs) to holistically assess long interleaved responses~\citep{opening,interleavedeval}, which is inherently misaligned with the fine-grained evaluation required to characterize complex generative capabilities~\citep{vbench}. Recent studies improve evaluation granularity through modality-specific decomposition or structured verification~\citep{ueval}. For example, ISG-Bench~\citep{isg} decomposes responses into four levels of structural granularity, but this decomposition does not map onto distinct capability dimensions. Moreover, it relies solely on a VLM evaluator across all levels, which may lack the specialized competence some dimensions require.

To address these limitations, we introduce IntReal, a real-world dataset and benchmark tailored to multimodal interleaved generation. We first identify four core generation scenarios from real-world interleaved content and organize them into a taxonomy. Guided by this taxonomy, we apply both understanding-oriented and generation-oriented filtering to a million-scale pool of raw interleaved data, resulting in \textbf{IntReal-100K}, a high-quality real-world dataset containing 538K images. Building on IntReal-100K, we further develop the \textbf{IntReal Evaluation Suite}, an evaluation framework grounded in real-world scenarios. Specifically, we construct an evaluation set through stringent automated filtering and expert curation. We then evaluate interleaved generation in two complementary settings: (1) holistic generation of complete interleaved responses to assess open-ended capabilities and (2) next-image generation conditioned on interleaved contexts for controlled diagnosis. Across both settings, we integrate specialized expert evaluators, such as image quality assessment (IQA) models, into seven metrics, enabling fine-grained diagnosis from diverse perspectives.

We conduct extensive experiments to assess IntReal Evaluation Suite and the utility of IntReal-100K for training UMMs. First, we examine three properties of the suite. Its seven metrics show low redundancy (maximum pairwise absolute Spearman correlation below 0.42), suggesting that they capture distinct capability dimensions. Automatic evaluations agree with human judgments in 83.3\% of cases, while results remain consistent across different expert evaluators (Spearman correlation above 0.9).  Second, supervised fine-tuning (SFT) on IntReal-100K improves text generation in models such as SenseNova-U1~\citep{sensenova-u1}, but degrades aspects of image generation, echoing the well-known tension between optimizing text and image generation in UMMs. Given the limited exploration of reinforcement learning for interleaved generation, we apply Flow-GRPO~\citep{flowgrpo} to next-image generation conditioned on interleaved contexts, while testing whether our metrics can serve as effective reward signals. Flow-GRPO recovers the degraded image generation capability and achieves the highest overall score among the open-source UMMs we evaluate on IntReal Evaluation Suite. These results show that training a single generation modality can improve interleaved generation, opening up optimization paths beyond conventional SFT on multimodal interleaved data. Together, these experiments show that IntReal-100K supports effective training of UMMs for interleaved generation, while IntReal Evaluation Suite accurately and reliably identifies strengths and weaknesses across multiple capability dimensions. 

Overall, our main contributions are summarized as follows:
\begin{enumerate}[leftmargin=*]
    \item \textbf{Real-world dataset.}
    We introduce IntReal-100K, a taxonomy-guided dataset for multimodal interleaved generation containing 100K high-quality, real-world interleaved sequences. Experiments demonstrate its utility for improving the interleaved generation capabilities of UMMs.
    \item \textbf{Fine-grained data-driven benchmark.}
    We develop IntReal Evaluation Suite, a data-driven benchmark that assesses interleaved generation across seven capability dimensions through coarse-to-fine evaluation. Three validation studies show that its metrics have low redundancy, align well with human judgments, and remain consistent across different expert evaluators.
    \item \textbf{Modality-specific gains transfer to interleaved generation.}
    We show that optimizing image generation under interleaved contexts can improve overall multimodal interleaved generation capability. This finding expands the optimization space beyond joint SFT on interleaved sequences.
\end{enumerate}

\section{Real-World Interleaved Dataset Construction}
\subsection{Characterizing Real-World Interleaved Content}

Web pages and videos are among the most abundant sources of naturally interleaved multimodal content. We collect a million-scale raw corpus of interleaved text-image content from seven sources that permit public collection and redistribution. Based on the functional roles and collaboration patterns between text and image, we analyze the collected data and identify four categories:
\begin{itemize}[leftmargin=*]
    \item \textbf{Planning:} \textit{Decision making under user-specific goals and constraints.} Text provides detailed information under given objectives, preferences, and constraints, while images visualize the corresponding spatio-temporal changes. We collect content from \textit{WikiVoyage} and \textit{WikiTravel}.
    \item \textbf{Knowledge:} \textit{Explaining concepts through real-world knowledge.} Text conveys concepts and explanations, while images provide corresponding visual representations and help ground abstract knowledge. We collect such encyclopedia content from \textit{Wikipedia}.
    \item \textbf{Tutorial:} \textit{Executing tasks through temporally ordered instructions.} Text describes procedural steps, while images illustrate intermediate states and temporal transitions to align instructions with execution, such as cooking tutorials. This category includes data collected from \textit{Xiachufang}.
    \item \textbf{Story:} \textit{Expressing ideas through narrative-driven content creation.} Text drives narrative progression, while images construct characters and scenes. Inspired by SEED-Story~\citep{seed-story}, this category includes data collected from \textit{The Land Before Time}, \textit{Rabbids}, and \textit{Curious George}.
\end{itemize}
As illustrated in~\cref{fig:teaser}, we visualize representative examples from the four representative categories identified above. Licensing details for each source are discussed in Appendix~\ref{app:data-source-and-licensing}.

\subsection{Data Collection and Curation}

High-quality training data for interleaved generation requires both individual modality quality and strong cross-modal alignment. To this end, we design a multi-stage curation pipeline that progressively improves data quality from low-cost filtering to high-cost semantic polishing, producing interleaved data for training UMMs. Specifically, we first apply low-cost visual quality filters to remove noisy and low-quality samples. We then leverage VLMs to polish textual descriptions while preserving their original factual content, improving readability and cross-modal consistency.  The overall pipeline consists of the following stages (see Appendix~\ref{app:detailed-filtering-pipeline} for detailed filtering criteria):

\paragraph{Step 1: Metadata-based Filtering.} We first filter images by file metadata, removing those with abnormal aspect ratios, low resolution, or unsupported formats to ensure basic visual quality.

\paragraph{Step 2: Representation-based Filtering.}
We employ representation-based specialized visual models to identify and remove undesirable content. Specifically, we use the RALI~\citep{rali} Image Quality Assessment (IQA) model to filter low-quality images, and finetune SigLIP2~\citep{siglip2} classifiers on expert-annotated in-domain data to detect watermarks and human faces. We further perform intra-sequence deduplication using DINOv3~\citep{dinov3} to remove near-duplicate images, ensuring both visual quality and diversity within each sample.

\paragraph{Step 3: VLM-based Refinement.} We employ Qwen3.5-397B-A17B~\citep{qwen3p5} to generate image captions and polish the surrounding text, following a principle that prioritizes original text over generated captions to preserve factual authenticity while improving text-image consistency.

\paragraph{Step 4: LLM-based Annotation.} We use Qwen3.5-397B-A17B to generate prompts and reasoning processes conditioned on the original content and its category, transforming each sample into structured multimodal interleaved data that can be directly used to fine-tune UMMs.

After the above curation process, we perform category-balanced sampling to ensure a basic balance across four categories. Only about 10\% of the collected samples are ultimately retained, as any defective segment within a long interleaved sample leads to its removal. This process results in a high-quality interleaved dataset, IntReal-100K, with category statistics provided in Appendix~\ref{app:dataset-statistic}.
\label{sec:2-data}

\section{Real-World Data-Driven Evaluation}
\begin{figure*}[t!]
  \centering
   \includegraphics[width=\linewidth]{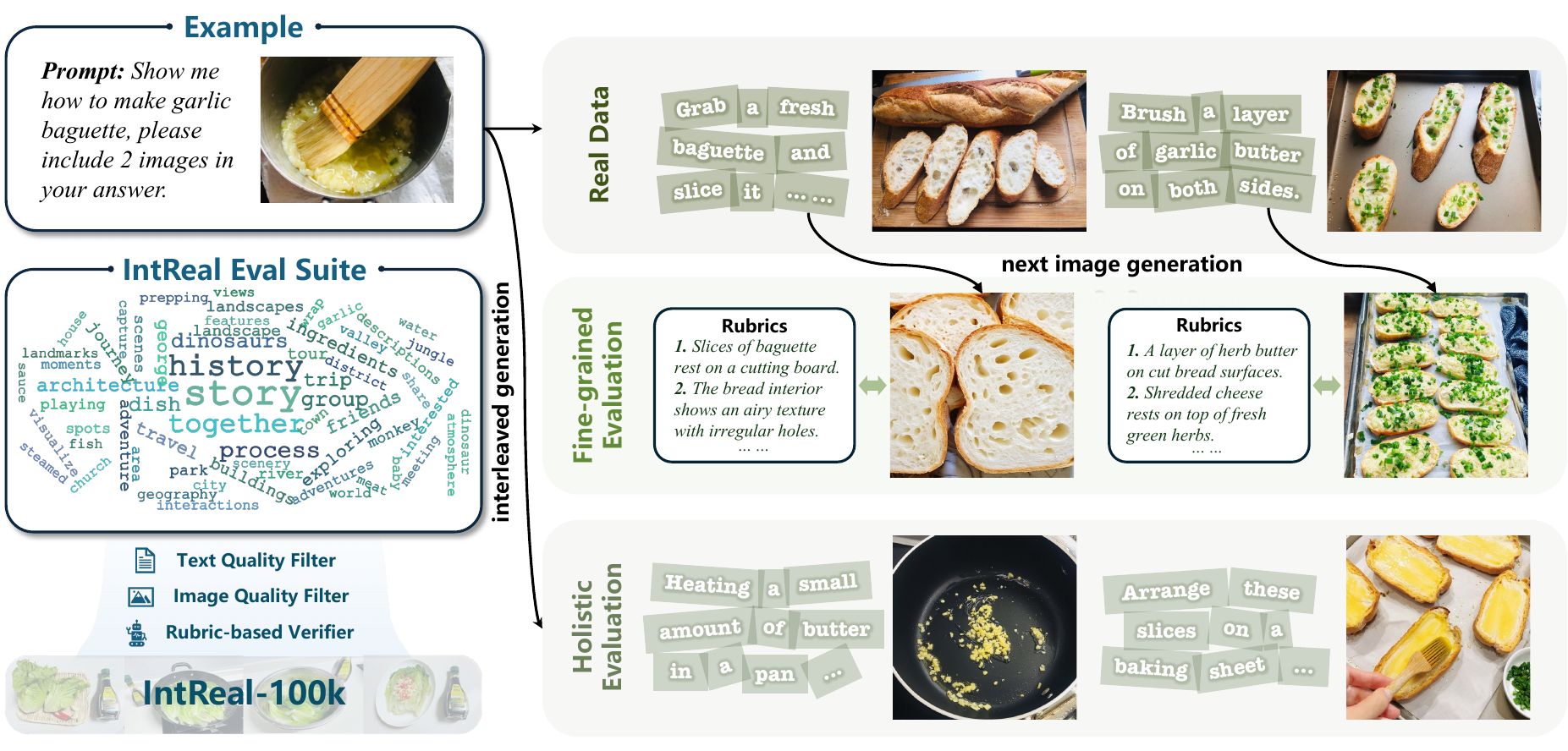}
   \caption{\textbf{Overview of the IntReal Evaluation Suite.} We select 600 high-quality samples from IntReal-100K to construct this suite. Each sample is evaluated at two granularities: holistic evaluation generates the complete interleaved sequence at once, while fine-grained evaluation generates the next image conditioned on the preceding interleaved context to enable controllable diagnosis.}
   \label{fig:eval}
\end{figure*}
As shown in~\cref{fig:eval}, we build our evaluation suite on real-world interleaved samples from IntReal-100K, combining a coarse-to-fine protocol with expert-guided, multi-dimensional assessment.
\subsection{Evaluation Set Construction}
\label{sec:eval-set}
To construct a reliable evaluation set, we adopt a semi-automatic curation strategy to select high-quality samples from IntReal-100K (see Appendix~\ref{app:evaluation-set-construction} for details). Specifically, we first build an automatic candidate pool by evaluating three key dimensions of interleaved sample quality: 

\begin{itemize}[leftmargin=*]
    \item \textbf{Text Quality:} We leverage Qwen3.5-397B-A17B to assess textual quality from three perspectives: prompt specificity, response richness, and topic informativeness.
    \item \textbf{Image Quality:} We use RALI to retain samples with high-quality images.
    \item \textbf{Text-Image Alignment:}  We adopt a context-aware self-verification strategy using Qwen3.5-397B-A17B. For each image, the model first derives visual rubrics specifying what the image should depict from the preceding text-image context, without accessing the target image. It then checks the target image against rubrics to filter out samples with cross-modal inconsistencies.
\end{itemize}

We adjust the filtering thresholds to obtain a high-quality candidate pool of manageable size, then manually remove samples that fail to meet the quality criteria. Finally, we perform stratified sampling by category to construct a balanced evaluation set of 600 samples, with 150 per category. We opt for a relatively compact evaluation set, since our coarse-to-fine protocol already requires generating over 6,000 images interleaved with text per model. Scaling further would substantially increase evaluation cost across multiple model families without commensurate benefit.

\subsection{Expert Model-based Coarse-to-Fine Benchmarking}
\label{sec:metric}
To evaluate UMMs in a coarse-to-fine manner, we design two complementary settings for interleaved generation. Holistic evaluation assesses complete interleaved responses generated from scratch. Fine-grained evaluation tests next-image generation under fixed real-world interleaved contexts. These shared contexts enable controlled, fine-grained comparisons across models. The two settings are illustrated in~\cref{fig:eval}. We design seven metrics to assess distinct capabilities of interleaved generation, where H and F denote metrics used for holistic and fine-grained evaluation, respectively:
\begin{itemize}[leftmargin=*]
    \item \textbf{Text Richness (H):} We employ an LLM as a text modality expert to compare generated textual responses with corresponding real-world textual responses through pairwise ranking, evaluating the richness and informativeness of generated texts.
    \item \textbf{Text Logicality (H):} Using the same LLM-based text expert, we assess whether generated texts contain factual errors, contradictions, or logical inconsistencies.
    \item \textbf{Image Aesthetic (H):} We employ an IQA model as a visual modality expert to measure the average aesthetic quality of generated images.
    \item \textbf{Image Diversity (H):} We identify visual repetition as a common failure mode, where images generated at consecutive positions may become overly similar and fail to reflect the progression of the context. To quantify this issue, we compute the DINO feature similarity between adjacent generated images, where extremely high similarity indicates potential visual repetition.
    \item \textbf{Quantity Compliance (H):} We use the number of images in the corresponding real-world sample as the target quantity and require the model to generate the same number, thereby evaluating instruction following with an image budget known to be sufficient for conveying the content.  
    \item \textbf{Context Fidelity (F):} We assess whether generated next images are consistent with their preceding interleaved context. We employ a VLM as a multimodal expert to verify whether each generated image satisfies the visual rubrics constructed during evaluation set curation in~\cref{sec:eval-set}.
    \item \textbf{Subject Consistency (F):} We compute DINO similarity between generated image and their corresponding reference image at the same positions in interleaved samples to measure whether key visual subjects remain consistent, e.g., story characters or cooking ingredients.
\end{itemize}
Detailed formulations of metrics are provided in Appendix~\ref{app:metric-definitions}. All metrics are evaluated across all categories, except Subject Consistency, which is computed only for Tutorial and Story. This is because Knowledge and Planning do not necessarily require consistent subjects across images.
\label{sec:3-eval}

\section{Experiment}
\begin{table*}[t!]
\centering
\caption{
\textbf{IntReal evaluation result per dimension}. We evaluate four UMM families across seven dimensions. All scores are scaled to the range of \([0,100]\), with higher values indicating better performance. The $\Delta$ vs.\ base rows report relative changes after post-training with respect to the corresponding base model, where green and red indicate improvements and degradations, respectively. Gray rows denote the closed-source API system, and $\Delta$ vs.\ OSS Best reports its relative difference with respect to the best open-source result in each dimension.
}
\resizebox{\textwidth}{!}{
\begin{tabular}{l|ccccc|cc|c}
\toprule
\multicolumn{1}{c|}{\multirow{3}{*}{\textbf{Model}}}
& \multicolumn{5}{c|}{\textbf{Holistic Evaluation}}
& \multicolumn{2}{c|}{\textbf{Fine-grained Evaluation}}
& \multirow{3}{*}{\textbf{Mean}} \\
\cmidrule(lr){2-6} \cmidrule(lr){7-8}
& \makecell[c]{Text\\Richness} & \makecell[c]{Text\\Logicality} & \makecell[c]{Image\\Aesthetic} & \makecell[c]{Image\\Diversity} & \makecell[c]{Quantity\\Compliance} & \makecell[c]{Context\\Fidelity} & \makecell[c]{Subject\\Consistency} & \\
\midrule
Emu3.5 & 14.02 & \textbf{75.17} & 81.20 & 75.51 & 39.70 & 65.60 & 61.46 & 58.95 \\
\midrule
U1-A3B-MoT & 24.67 & 69.17 & 80.70 & 88.70 & 72.50 & 86.81 & 63.45 & 69.43 \\
\midrule
Bagel\textit{+SFT} & 17.50 & 63.50 & 76.82 & 91.35 & 62.83 & 74.19 & 60.61 & 63.83 \\
Bagel\textit{+SFT+RL} & 16.67 & 62.50 & \textbf{81.94} & 90.31 & 62.50 & 76.04 & 65.43 & 65.06 \\
\midrule
U1-8B-MoT & 21.83 & 67.33 & 79.60 & 88.36 & \textbf{95.33} & \textbf{87.29} & 63.38 & 71.88 \\
\quad\textit{+SFT} & \textbf{28.17} & 71.83 & 75.10 & 87.05 & 93.50 & 83.02 & \textbf{67.52} & 72.31 \\
\(\Delta\) vs. base (\%) 
& \cellcolor{DeltaGreen!100.00}\textit{+29.04\%}
& \cellcolor{DeltaGreen!66.8}\textit{+6.68\%}
& \cellcolor{DeltaRed!56.5}\textit{\textminus5.65\%}
& \cellcolor{DeltaRed!14.8}\textit{\textminus1.48\%}
& \cellcolor{DeltaRed!19.2}\textit{\textminus1.92\%}
& \cellcolor{DeltaRed!48.9}\textit{\textminus4.89\%}
& \cellcolor{DeltaGreen!65.3}\textit{+6.53\%}
& \cellcolor{DeltaGreen!6.0}\textit{+0.60\%} \\
\quad\textit{+SFT+RL} & 27.83 & 70.50 & 78.53 & \textbf{93.31} & 93.83 & 85.57 & 67.11 & \textbf{73.81} \\
\(\Delta\) vs. base (\%)
& \cellcolor{DeltaGreen!100.00}\textit{+27.49\%}
& \cellcolor{DeltaGreen!47.1}\textit{+4.71\%}
& \cellcolor{DeltaRed!13.4}\textit{\textminus1.34\%}
& \cellcolor{DeltaGreen!56.0}\textit{+5.60\%}
& \cellcolor{DeltaRed!15.7}\textit{\textminus1.57\%}
& \cellcolor{DeltaRed!19.7}\textit{\textminus1.97\%}
& \cellcolor{DeltaGreen!58.9}\textit{+5.89\%}
& \cellcolor{DeltaGreen!26.9}\textit{+2.69\%} \\
\midrule
\rowcolor{gray!7}
NanoBanana2 & 90.80 & 97.49 & 82.32 & 83.56 & 95.32 & 76.70 & 67.65 & 84.84 \\
\rowcolor{gray!7}
$\Delta$ vs. OSS Best (\%)
& +222.33\%
& +29.69\%
& +0.46\%
& -10.45\%
& -0.01\%
& -12.13\%
& +0.19\%
& +14.94\% \\
\bottomrule
\end{tabular}
}
\label{tab:main-results}
\end{table*}

\subsection{Experimental Setup}
\label{sec:exp-setup}
We evaluate four families of UMMs with varying levels of support for interleaved generation and fine-tuning: \textbf{(1) Bagel}~\citep{bagel}, which provides official fine-tuning scripts but whose released checkpoint does not natively support interleaved generation. We therefore report only its performance after fine-tuning on IntReal-100K. \textbf{(2) Emu3.5}~\citep{emu3p5}, which supports native interleaved inference but does not release fine-tuning methods. We thus evaluate only its original checkpoint. \textbf{(3) SenseNova-U1}~\citep{sensenova-u1}, which supports both fine-tuning and interleaved inference, allowing evaluation before and after fine-tuning on IntReal-100K. SenseNova-U1 provides two variants, 8B-MoT and A3B-MoT. Both use an 8B generation branch, while their understanding components adopt an 8B and a 30B-A3B MoE architecture, respectively. For 8B-MoT, the authors additionally release \textit{SenseNova-U1-8B-MoT-Interleaved}, a checkpoint specifically optimized for this task. To ensure a strong baseline, we adopt this checkpoint for our evaluation and finetuning. No corresponding specialized checkpoint is released for A3B-MoT, for which we therefore evaluate the standard model. Due to computational constraints, we report fine-tuned results only for the 8B-MoT variant. \textbf{(4) NanoBanana2 (Gemini3.1-Flash-Image)}, which is a closed-source API supporting interleaved generation. Since the system is not publicly disclosed, we cannot determine whether this capability is native or implemented through an agentic pipeline. Other models are excluded if they are unavailable or lack publicly released implementations required for our finetuning and evaluation such as LLaDA2.0-Uni~\citep{llada2uni}. In addition, Anole~\citep{anole} and Show-o2~\citep{showo2} are excluded from our evaluation because their released models are limited in context length and cannot support the long-context interleaved generation setting.

For evaluation, we use Gemma4~\citep{gemma4} for both LLM- and VLM-based judging due to its strong performance, open-source availability, and modest computational requirements. For fine-tuning, we use the official training framework of each model for SFT, while all RL experiments are implemented with Flow-Factory~\citep{flowfactory}. All post-training data are drawn from the 99.4K samples in IntReal-100K after excluding the 600 samples used in the IntReal Evaluation Suite, ensuring that the training and evaluation sets are strictly disjoint. Additional evaluation details are provided in Appendix~\ref{sec:app-bench}, and further training and inference details are provided in Appendix~\ref{sec:app-exp}.

\begin{figure}[t]
    \centering
    \begin{minipage}[t]{0.51\linewidth}
        \centering
        \adjustbox{valign=t}{\includegraphics[width=\linewidth]{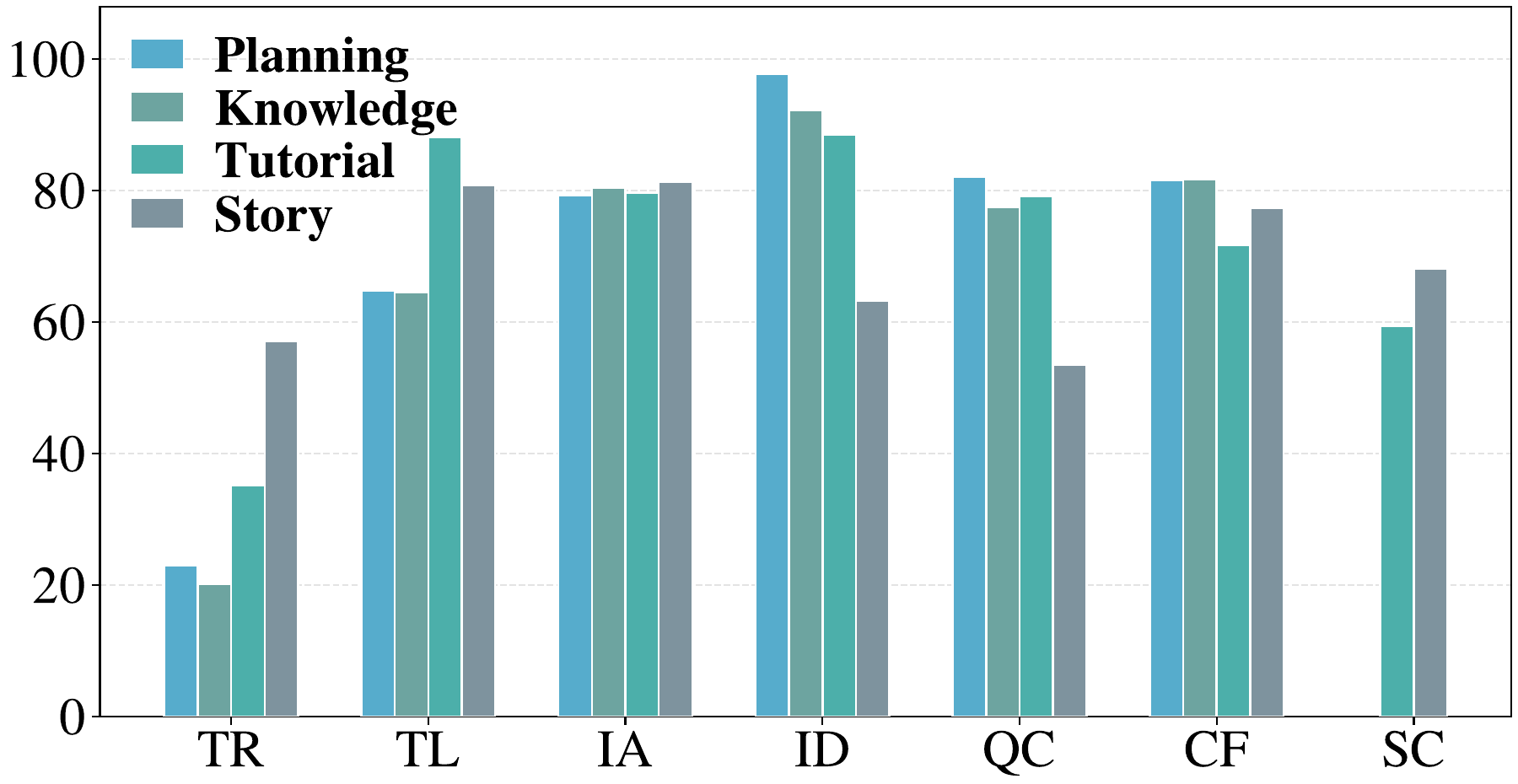}}
        \caption{\textbf{IntReal evaluation results per category.} We report the score of each metric for each category, computed by averaging over UMMs.}
        \label{fig:per-cat}
    \end{minipage}
    \hfill
    \begin{minipage}[t]{0.45\linewidth}
        \centering
        \adjustbox{valign=t}{\includegraphics[width=\linewidth]{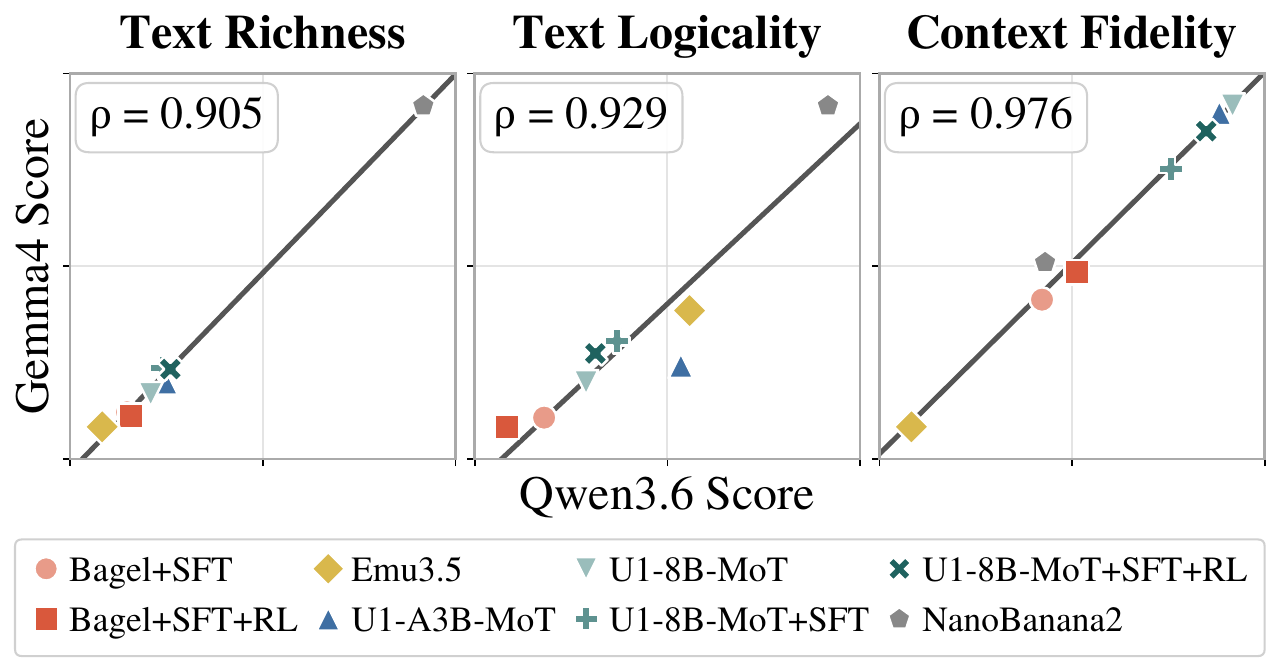}}
        \caption{\textbf{Spearman correlations across metrics with different VLM evaluators.} The consistent model rankings demonstrate the robustness of our evaluation method.}
        \label{fig:metric-robustness}
    \end{minipage}
\end{figure}

\subsection{Benchmarking UMMs on IntReal}
\Cref{tab:main-results} presents the per-dimension IntReal evaluation results, while~\cref{fig:per-cat} further reports the results across different categories. From these results, we make the following observations.

\paragraph{Different UMMs exhibit distinct and complementary capabilities.}
No single model performs consistently well across all evaluation dimensions. Among open-source models, Emu3.5 achieves the strongest Text Logicality, while U1-8B-MoT performs substantially better in Quantity Compliance and Context Fidelity. The closed-source NanoBanana2 also does not uniformly outperform open-source models, showing advantages on unimodal dimensions such as Text Richness, Text Logicality, and Image Aesthetic, while remaining below the best open-source results on interleaved-specific dimensions such as Image Diversity and Context Fidelity. These results suggest that strong unimodal capabilities are important for interleaved generation, but are not sufficient on their own, as interleaved generation also requires optimization for interleaved-specific dimensions.

\paragraph{Different categories emphasize distinct capabilities.}
Planning and Knowledge are more demanding in Text Richness and Text Logicality, while Tutorial is particularly challenging under fine-grained evaluation, with lower Context Fidelity and Subject Consistency. Story often requires longer interleaved sequences, leading to greater challenges in Quantity Compliance and Image Diversity.

\paragraph{OSS UMMs exhibit imbalanced multimodal capabilities.}
The largest gap between open-source UMMs and the closed-source API lies in the text modality, whereas the gap in image modality is substantially smaller. This asymmetry is consistent with recent findings on scaling UMMs~\citep{beyond-llms,ernie5}, where language is more parameter-hungry while vision is more data-hungry. Given that current OSS UMMs remain much smaller than today's flagship LLMs, scaling model capacity may be one of the most important steps toward stronger interleaved generation.

\subsection{Post-training on IntReal-100K}

Compared with the original U1-8B-MoT, SFT improves Text Richness, Text Logicality, and Subject Consistency by 29.04\%, 6.68\%, and 6.53\%. However, Image Aesthetic and Context Fidelity decrease by 5.65\% and 4.89\%, revealing a trade-off between different capabilities. This degradation may arise from the inherent optimization conflict between understanding and generation in UMMs. We therefore investigate whether Reinforcement Learning (RL) can alleviate this trade-off.

\begin{figure*}[t!]
  \centering
   \includegraphics[width=\linewidth]{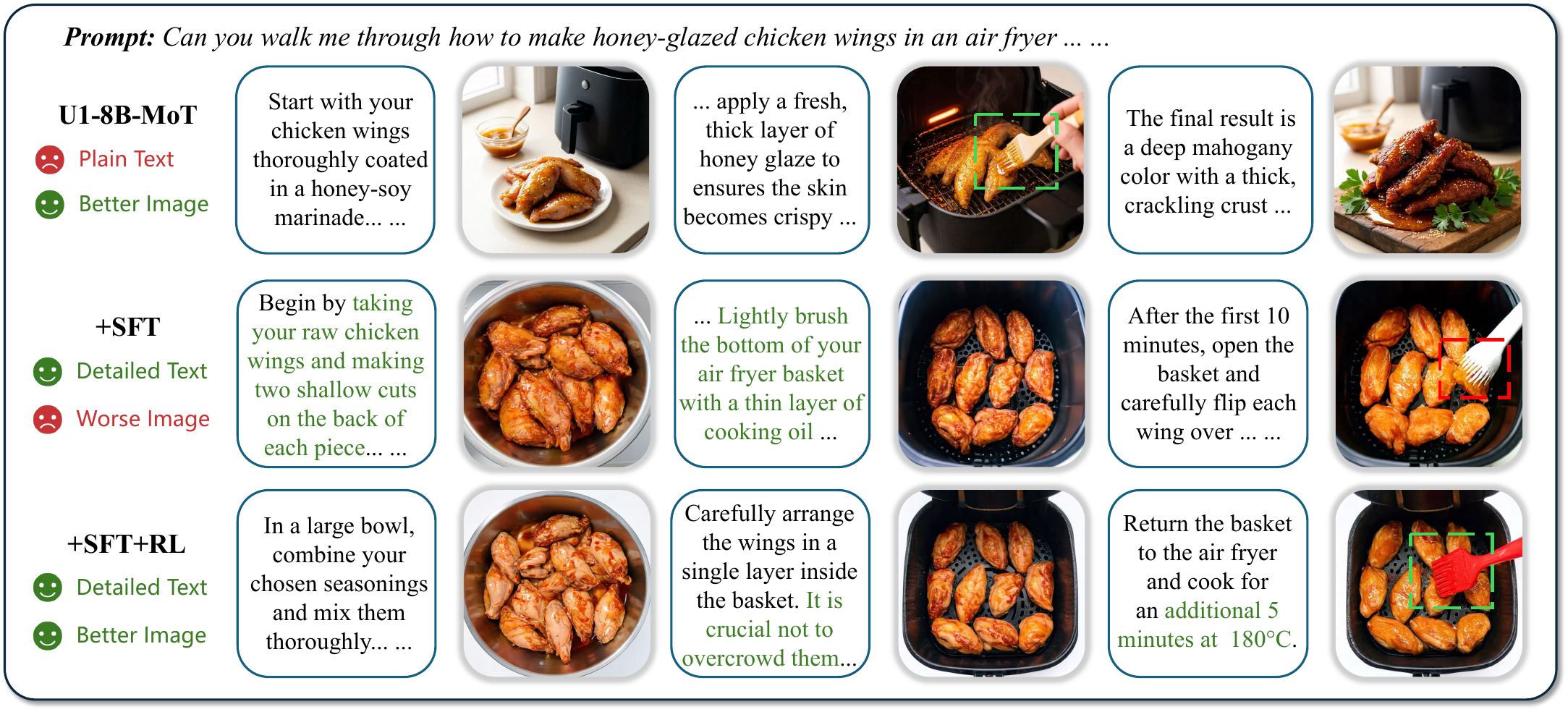}
   \caption{\textbf{Case study on SenseNova-U1-8B-MoT.} SFT produces more detailed text but degrades visual details (distorted brush), while RL preserves the text improvements and restores image quality.}
   \label{fig:case-study}
\end{figure*}

To close the loop, we validate whether the metrics introduced in~\cref{sec:metric} can serve as effective learning signals. We first select 10K high-quality RL samples and annotate them with rubrics following the same procedure in~\cref{sec:eval-set}. We then apply FlowGRPO~\citep{flowgrpo} using IQA, DINO, and rubric-based rewards for optimization. For U1-8B-MoT, RL substantially mitigates the degradation in Image Aesthetic and Context Fidelity introduced by SFT, while further improving Image Diversity. A similar trend holds for Bagel before and after RL in Image Aesthetic and Context Fidelity, while its already-high Image Diversity remains stable. Overall, U1-8B-MoT with SFT and RL achieves the highest mean score among all OSS models, indicating that improvements in image-generation capability can effectively transfer to interleaved generation. A case study of U1-8B-MoT under different post-training settings is shown in~\cref{fig:case-study}.

\begin{figure}[t]
    \centering
    \begin{minipage}[t]{0.53\linewidth}
        \centering
        \adjustbox{valign=t}{\includegraphics[width=\linewidth]{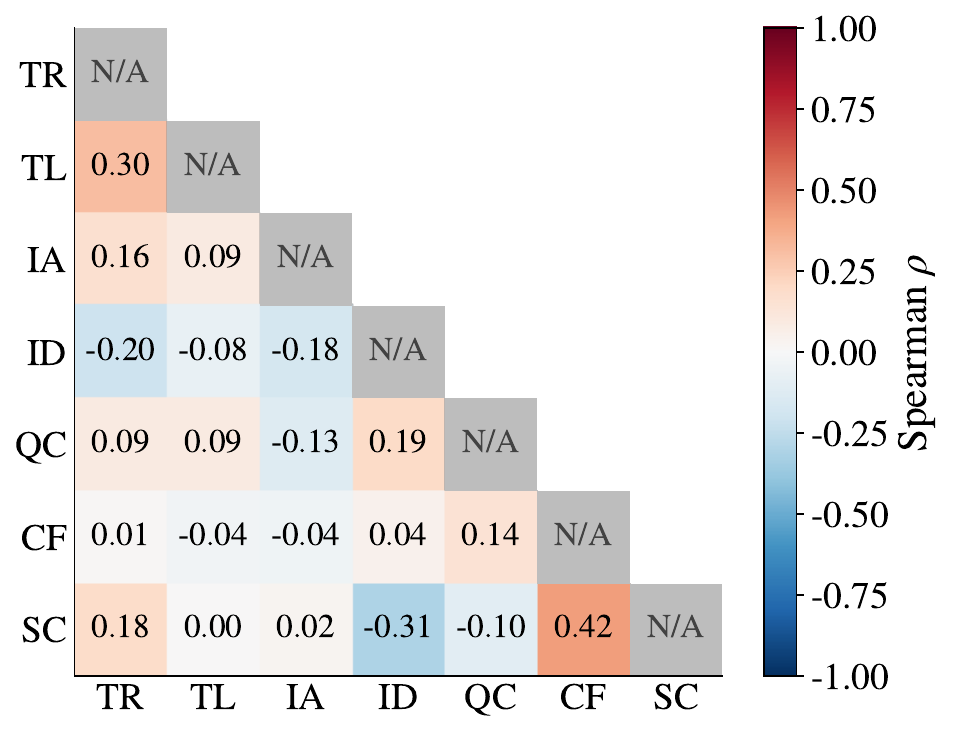}}
        \caption{\textbf{Pairwise sample-level Spearman correlation between different evaluation metrics.} Low absolute correlations indicate limited redundancy.}
        \label{fig:intra}
    \end{minipage}
    \hfill
    \begin{minipage}[t]{0.43\linewidth}
        \centering
        \adjustbox{valign=t}{\includegraphics[width=\linewidth]{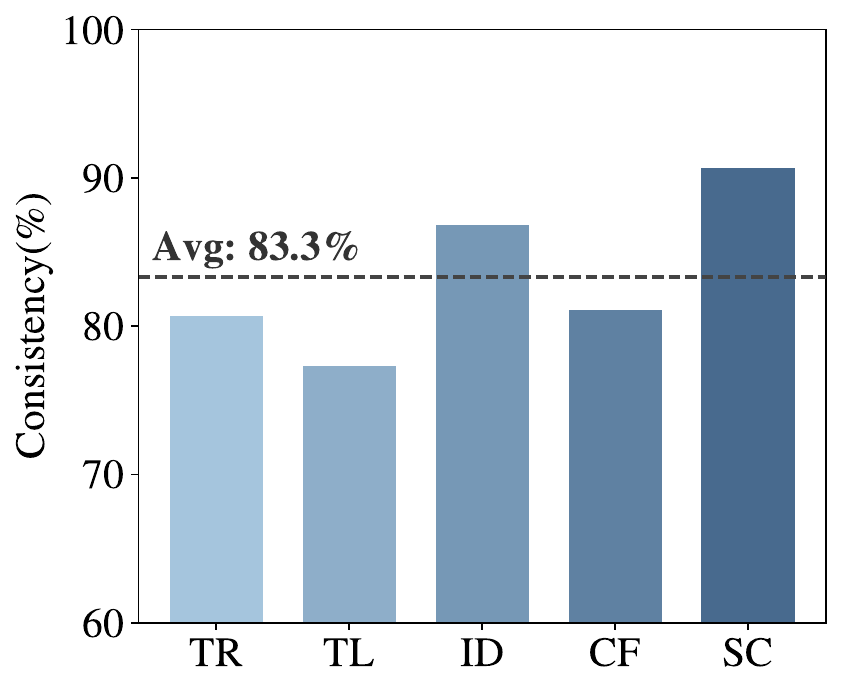}}
        \caption{\textbf{Agreement between automatic and human annotation for metrics.} Metrics with direct measurements (Quantity Compliance) or human-aligned evaluation models (Image Aesthetic) are excluded.}
        \label{fig:human}
    \end{minipage}
\end{figure}

\subsection{Reliability of the IntReal Evaluation}
\label{sec:exp-reliability}
To validate the reliability of our proposed evaluation method, we conduct three complementary analyses. We select Emu3.5, U1-A3B-MoT, Bagel-SFT, U1-8B-MoT and NanoBanana2, avoiding additional fine-tuned variants to prevent repeatedly evaluating highly similar models.

\paragraph{Metric Orthogonality.} We first investigate the numerical redundancy and orthogonality among the seven metrics. For each selected model, we compute pairwise sample-level Spearman rank correlations between all metric pairs and average the correlations across models. As shown in~\cref{fig:intra}, the absolute Spearman correlation between any pair of metrics remains below 0.42, indicating that the proposed metrics provide complementary measurements with limited redundancy.

\paragraph{Human-Automatic Agreement.}
We focus on five subjective metrics, excluding Image Aesthetic and Quantity Compliance, since the former is evaluated by a human-aligned IQA model and the latter is objectively computed. For representative models, we construct all $\binom{5}{2}=10$ model pairs and sample 15 comparisons for each pair, resulting in 150 pairwise comparisons per metric and 750 expert-annotated comparisons in total. The automatic evaluation agrees with expert judgments in 83.3\% of comparisons on average, demonstrating strong human-automatic consistency.

\paragraph{Robustness to expert evaluators.}
We assess the robustness of our VLM-based metrics by comparing Gemma4 with Qwen3.6-35B-A3B~\citep{qwen3p6}, an open-source model of similar scale. For Text Richness, Text Logicality, and Context Fidelity, we compute the Spearman correlation between all model rankings produced by the two evaluators. As shown in~\cref{fig:metric-robustness}, all correlations exceed 0.9, indicating that our metrics maintain stable model rankings across different evaluators.
\label{sec:4-exp}

\section{Related Work}
\paragraph{Unified Multimodal Models (UMMs).}
Existing research on UMMs has primarily explored two complementary directions. The first seeks to unify text and image generation within a single model~\citep{dalle,emu,janus,transfusion,blip3o,cheers,mmada,dreamllm,longcat-next,omnidiffusion}. Representative studies combine multimodal understanding with text-to-image generation, while subsequent works further incorporate image editing and other any-to-image tasks to enable image generation conditioned on multimodal contexts~\citep{showo,vilau,tuna,hunyuanimage3}. The second treats interleaved text-image sequences as a native form of multimodal pretraining rather than a collection of separate downstream tasks~\citep{emu3p5,beyond-llms,physics-mm,mm-interleave,orthus,llada2uni}. Chameleon~\citep{chameleon} exemplifies this formulation through mixed-modal sequence modeling, while Bagel~\citep{bagel} provide s a large-scale open-source realization based on interleaved multimodal data. 

\paragraph{Multimodal Interleaved Datasets and Benchmarks.}
Early interleaved datasets were primarily developed for multimodal understanding and pretraining, constructed through text-based synthesis or web-scale collection~\citep{mmc4,obelics,omnicorpus,mint}. CoMM~\citep{comm} further introduces generation-oriented filtering, making interleaved data more suitable for generation tasks. Recent studies have shifted toward enabling models to learn multimodal reasoning traces~\citep{thinkmorph,zebracot}. On the evaluation side, early benchmarks mainly define diverse downstream tasks for interleaved generation and commonly rely on VLMs for holistic assessment~\citep{opening,interleavedeval}. UEval~\citep{ueval} and ISG-Bench~\citep{isg} take steps toward fine-grained evaluation by decoupling interleaved responses into more structured evaluation targets. Meanwhile, Uni-MMMU~\citep{unimmmu} and RealUnify~\citep{realunify} benchmark the bidirectional synergy between understanding and generation through tasks that explicitly couple textual reasoning and visual generation. In contrast, WEAVE~\citep{weave} evaluates UMMs over multi-turn interleaved text-image contexts, with each turn producing a single-modality response. Beyond UMMs, UniM~\citep{unim} evaluates omni-modal models under arbitrary combinations of inputs and outputs across multiple modalities.

\label{sec:5-relate}

\section{Conclusion}
In this work, we systematically study multimodal interleaved generation from the perspectives of application scenarios, data construction, evaluation, and post-training. We first characterize four representative real-world scenarios and develop a taxonomy-guided data curation pipeline to construct IntReal-100K from real-world content. Building upon these data, we introduce the IntReal Evaluation Suite, which follows a coarse-to-fine evaluation principle and integrates specialized expert models to provide complementary measurements across seven dimensions. Our experiments reveal that SFT on IntReal-100K brings gains but also introduces trade-offs across different capabilities. Through Flow-GRPO, we further show that the proposed metrics can serve as effective learning signals, and that improving unimodal generation capability can generalize to multimodal interleaved generation. These findings suggest that modality-specific optimization can improve interleaved generation, expanding the training options beyond joint optimization of text and image generation. IntReal also has several limitations. Constructing high-quality real-world interleaved data requires aggressive filtering and thus yields a relatively low retention rate. In addition, we do not explore mixed post-training with unimodal generation data to maintain a clean experimental setting. We leave synthetic data augmentation and mixed-data post-training for future work.
\label{sec:con}

\bibliography{iclr2027_conference}
\bibliographystyle{iclr2027_conference}

\clearpage
\appendix

\startcontents[appendix]

\begin{center}
  {\Large\bfseries Appendix}
\end{center}

\begingroup
  \hypersetup{hidelinks}
  \printcontents[appendix]{}{1}{\setcounter{tocdepth}{2}}
\endgroup

\bigskip

\section{Dataset Details}
\subsection{Data Sources and Licensing}
\label{app:data-source-and-licensing}
\paragraph{Wikivoyage.}
Wikivoyage is a collaborative, community-edited travel guide hosted by the Wikimedia Foundation.  It provides destination-oriented articles covering attractions, transportation, accommodation, local customs, and suggested itineraries, making it a natural source of planning-oriented interleaved content. Textual content on Wikivoyage is released under the Creative Commons Attribution-ShareAlike license, which permits copying, redistribution, and adaptation provided that appropriate attribution is given and derivative works are distributed under the same or a compatible license. Images and other media may be subject to file-specific licenses, and we retain the corresponding provenance and licensing information when collecting the data.  Our use and release of Wikivoyage-derived content follow the applicable attribution and share-alike requirements.

\paragraph{Wikitravel.}
Wikitravel is a collaboratively written travel guide containing destination descriptions, and itinerary-oriented content contributed by its community. Its structured combination of descriptive text and travel-related images provides another source of planning-oriented multimodal sequences in IntReal. Wikitravel requires its textual contributions to be licensed under the Creative Commons Attribution-ShareAlike 3.0 license or a compatible license, allowing redistribution and modification with appropriate attribution and share-alike requirements. Uploaded images and other media are governed by their respective file-specific licenses. Therefore, we inspect the licensing metadata associated with each sample and filter out content whose licensing terms do not permit redistribution. All Wikitravel-derived data released in IntReal follow these licensing conditions.

\paragraph{Wikipedia.}
Wikipedia is a collaboratively maintained online encyclopedia hosted by the Wikimedia Foundation, covering a broad range of real-world concepts, entities, events, and scientific topics.  Its articles naturally integrate explanatory text with illustrative diagrams and photographs, making Wikipedia the primary source of knowledge-oriented interleaved content in IntReal. Wikipedia textual content is distributed under Creative Commons licenses that permit reuse, redistribution, and adaptation subject to attribution and share-alike requirements. Images and other media are individually licensed, including Creative Commons  and other compatible licenses, and their licensing information is provided on the corresponding file pages. During data construction, we preserve source and license metadata and only redistribute media whose individual licenses permit such use.

\paragraph{Xiachufang.}
Xiachufang is a community-based cooking platform where users share recipes and cooking instructions, providing naturally aligned procedural text-image sequences for the Tutorial in IntReal. According to Xiachufang's community guidelines, the copyright of recipes, images, and textual content published by users remains with the corresponding content creators. The platform permits third parties to reproduce user-generated content for non-commercial purposes provided that the original author and Xiachufang are clearly attributed, the original source link is retained. Such permission is subject to individual users' copyright declarations. During data curation, we inspect the available copyright and attribution metadata and filter out samples for which the user explicitly prohibits redistribution or whose licensing conditions are incompatible with our intended research use. For each retained image, we preserve its original source URL, and the corresponding original textual content is also retained for provenance and attribution. The VLM-polished text introduced during data curation is used only as training supervision. All retained data are used solely for non-commercial academic research.

\paragraph{Curious George.}
Curious George is an educational animated series centered on the adventures of George,
a curious monkey who explores everyday problems through observation, experimentation,
and problem solving. The series is designed to introduce young audiences to concepts in science, engineering, and mathematics, and is distributed through PBS KIDS. Curious George is a production of Imagine, WGBH, and Universal. The characters created
by Margret and H.A. Rey are copyrighted and trademarked by Houghton Mifflin Harcourt
and used under license from Universal Studios Licensing LLC. The corresponding PBS terms of use permit downloading materials for non-commercial, personal use with copyright notices retained, and additionally allow limited reproduction of website materials by schools and libraries for educational purposes.

\paragraph{Rabbids Invasion.}
Rabbids Invasion is an animated comedy series featuring the Rabbids, a group of
mischievous rabbit-like characters whose episodic adventures form short, visually driven stories. We use episodes from the series as a source of naturally occurring interleaved story content. The series is currently distributed by Animaj, whose official page lists Rabbids Invasion as a four-season children's animated series and identifies its role as distribution. The audiovisual content remains subject to the copyright and intellectual-property rights of its respective rights holders. We use the collected material exclusively for non-commercial academic research and retain provenance information linking each sample to its original source. 

\paragraph{The Land Before Time.} The Land Before Time is a long-running animated franchise following a group of young dinosaurs, including Littlefoot, Cera, Ducky, Petrie, and Spike, through narrative adventures centered on friendship, exploration, and cooperation. We derive story samples from episodes and films in the franchise to represent long-form narrative interleaved content. The franchise is distributed by Universal Pictures, and the corresponding official releases are copyrighted by Universal Studios with all rights reserved. We use the collected material only for non-commercial academic research and maintain the provenance of each derived sample. 

For the three animated series above, we follow the data organization and release practices of StoryStream~\citep{seed-story}, which has publicly released image–text samples derived from the same sources. We sincerely thank the creators, contributors, and maintainers of all data sources used in IntReal. The provenance and applicable licensing information for each source are described above. All data collected in IntReal are intended solely for non-commercial academic research, and we respect the intellectual property rights of the original content creators. For openly licensed content, we follow the corresponding license requirements, including attribution and redistribution conditions where applicable. Users of IntReal are responsible for complying with the copyright, licensing terms, and usage policies of the respective original data sources. 

\subsection{Detailed Filtering Pipeline}
\label{app:detailed-filtering-pipeline}

We apply a unified multi-stage filtering pipeline to all collected sources, while adapting several filtering criteria to the characteristics of individual domains. After filtering, we use VLMs to caption the remaining high-quality visual content, followed by LLM-based refinement and annotation grounded in the original real-world samples. This process improves text-image consistency and converts the collected content into trainable supervision while preserving its factual authenticity.

\paragraph{Filtering Rules.}
For metadata-based filtering, we retain images with aspect ratios between $1{:}2$ and $2{:}1$ and exclude unsupported formats, including GIF and WebP. We further impose source-specific minimum resolution requirements on the shorter side of each image. Specifically, we require a minimum shorter-side resolution of 512 pixels for Wikivoyage, Wikitravel, Wikipedia, and Xiachufang, while using a threshold of 384 pixels for the three story sources. The relaxed threshold for story data accounts for the relatively low native resolution of older animated content, whose shorter side frequently falls below 512 pixels.

For representation-based filtering, we adapt the activation of different filters according to the characteristics of each source. Since Xiachufang primarily consists of user-uploaded content, we apply both face and watermark filtering to remove potentially undesirable user-generated images. We fine-tune one SigLIP2-based~\citep{siglip2} binary classifier for these tasks using 5000 manually annotated Xiachufang samples, achieving an F1 score of 0.82 on the held-out test set. In contrast, we do not apply face filtering to the three Wiki-based sources, since human faces often constitute semantically necessary content, such as portraits of notable individuals, while these sources already undergo platform-level content moderation and rarely contain intrusive watermarks. Similarly, face filtering is not applied to the three story sources, where recurring characters constitute essential visual subjects and should be preserved.

Across all sources, we additionally apply RALI-based~\citep{rali} image quality filtering and DINO-based intra-sequence deduplication. For RALI, we require each individual image to obtain a score above 3.0 and the average score of all images within a sample to exceed 3.5. For DINO-based deduplication, adjacent images are required to have a feature similarity below 0.85, which removes highly repetitive visual sequences while preserving meaningful visual continuity.

\paragraph{Content Refinement and Annotation.}
After filtering, all retained samples undergo a four-stage refinement and annotation pipeline to improve text-image alignment while preserving the factual content of the original sources. First, we generate detailed captions for each image to recover visual information that may be absent from the surrounding text. Second, we refine the original textual content by incorporating complementary information from the generated captions, while explicitly preserving the facts and semantics of the source content. Third, we construct hierarchical annotations at both the sequence and image levels: for each interleaved sample, we annotate its user prompt and generate a sequence-level reasoning trace, while each individual image is additionally associated with an image-specific reasoning trace describing its role and generation rationale within the sequence. Finally, we jointly refine the complete prompt and interleaved content to further improve their global consistency and coherence. Through this process, the collected web content is transformed into training-ready interleaved data with improved multimodal alignment, structured supervision, and preserved factual fidelity.

\subsection{Dataset Statistics}
\label{app:dataset-statistic}
\Cref{tab:dataset-statistics} presents the statistics of IntReal-100K across different categories and data sources. Different categories exhibit distinct structural characteristics. Planning data are relatively concise, with fewer images and shorter text per sample, while Knowledge and Tutorial data contain richer multimodal content. Xiachufang constitutes the largest individual source, whereas Story samples generally contain the longest interleaved sequences with around eight images per sample. At the category level, the dataset maintains a relatively balanced coverage of diverse real-world content. Overall, IntReal-100K contains 100K samples, 538K images, and 146.6M text tokens.

\begin{table}[t]
\centering
\caption{
\textbf{Statistics of IntReal-100K across different data sources.}
We report the number of samples, number of images, average number of images per sample, total number of text tokens, and average number of text tokens per sample. Text tokens are computed using the Qwen2.5 tokenizer.
}
\resizebox{\linewidth}{!}{
\begin{tabular}{llrrrrr}
\toprule
\textbf{Category}
& \textbf{Source}
& \textbf{\# Samples}
& \textbf{\# Images}
& \textbf{Images / Sample}
& \textbf{\# Text Tokens}
& \textbf{Text Tokens / Sample} \\
\midrule
Planning
& Wikivoyage
& 21,055 & 56,527 & 2.68 & 21.5M & 1,021.13 \\
& Wikitravel
& 5,055 & 12,254 & 2.42 & 5.2M & 1,020.53 \\
\midrule
Knowledge
& Wikipedia
& 21,138 & 100,198 & 4.74 & 32.2M & 1,522.34 \\
\midrule
Tutorial
& Xiachufang
& 34,771 & 213,818 & 6.15 & 52.3M & 1,504.77 \\
\midrule
Story
& Curious George
& 7,298 & 63,088 & 8.64 & 14.2M & 1,941.43 \\
& Rabbids Invasion
& 3,949 & 34,284 & 8.68 & 7.8M & 1,965.36 \\
& The Land Before Time
& 6,734 & 58,283 & 8.66 & 13.5M & 2,007.22 \\
\midrule
\textbf{Total}
& \textbf{7 sources}
& \textbf{100,000}
& \textbf{538,452}
& \textbf{5.38}
& \textbf{146.6M}
& \textbf{1,466.06} \\
\bottomrule
\end{tabular}
}
\label{tab:dataset-statistics}
\end{table}

\section{Benchmark Details}
\label{sec:app-bench}
\subsection{Evaluation Set Construction}
\label{app:evaluation-set-construction}
\paragraph{Initial Filtering.}
We first apply several basic constraints to remove samples that are unsuitable for evaluation. Specifically, we retain samples with an average image aesthetic score of at least 3.8 and between 3 and 10 output images. This ensures sufficient visual quality while avoiding samples with too few images to reflect interleaved generation or excessively long sequences that may introduce disproportionate evaluation cost.

\paragraph{Content Quality Assessment.}
For the remaining samples, we use Qwen3.5-397B-A17B to assess content quality along three dimensions on a 1--5 scale: \emph{prompt specificity}, \emph{response richness}, and \emph{topic informativeness}. We retain samples with prompt specificity $\geq 3$, response richness $\geq 4$, and topic informativeness $\geq 4$. Prompt specificity ensures that the prompt provides a sufficiently clear generation goal, while response richness encourages outputs with substantive and informative content. The topic informativeness criterion is used to reduce highly idiosyncratic or sparse examples, such as unusual user-created recipes or obscure encyclopedic topics, which may be difficult to evaluate reliably.

\paragraph{Context-Aware Rubric Verification.}
We further verify whether each image is sufficiently supported by its preceding interleaved context. For each target image, a VLM generates a set of visual rubrics based only on the preceding context. We require each rubric to be atomic, focusing on a single verifiable visual property without combining multiple conditions, and self-contained, such that it can be understood without referring back to the original context. In particular, ambiguous pronouns or context-dependent entity references are avoided.

We then use Qwen3.5-397B-A17B to independently verify each rubric against the corresponding image. A sample is retained only when every image satisfies at least one generated rubric and the overall rubric pass rate across the sample is at least 50\%. We do not require all rubrics to be satisfied, since naturally occurring interleaved content may contain imperfect image--text correspondence. This relaxed criterion preserves meaningful contextual consistency while retaining a moderate level of difficulty in the evaluation set.

\paragraph{Human Validation and Sampling.}
The automatic filtering pipeline produces approximately 1.5K candidate samples. The authors then manually review these candidates according to the same criteria and remove remaining failure cases. Finally, we perform stratified sampling across the four categories and different output image counts to obtain a balanced evaluation set of 600 samples.

\subsection{Metric Definitions}
\label{app:metric-definitions}

We provide detailed definitions of the seven metrics used in IntReal. Unless otherwise specified, each metric is normalized to $[0,1]$ and multiplied by 100 for reporting, with the convention that higher values consistently indicate better performance across all metrics. We denote the model-generated and reference texts of a sample as $T^{m}$ and $T^{g}$, and the generated and reference images at position $i$ as $I_i^{m}$ and $I_i^{g}$, respectively.

\paragraph{Text Richness (TR).}
Text Richness measures whether the generated textual response is at least as informative and complete as the corresponding real-world response. An LLM compares $T^{m}$ and $T^{g}$ through pairwise ranking, considering informativeness, completeness, and unnecessary redundancy. To reduce positional bias, we alternate the order of the generated and reference responses during evaluation. We treat both a model win and a tie as successful, since the objective is to determine whether the generated text reaches the quality of real-world content:

\begin{equation}
\mathrm{TR}=
\begin{cases}
1, & T^{m}\text{ wins or ties},\\
0, & T^{g}\text{ wins}.
\end{cases}
\label{eq:TR}
\end{equation}

\paragraph{Text Logicality (TL).}
Text Logicality evaluates the internal coherence and factual plausibility of the generated interleaved response. The LLM checks whether the response follows the entities and constraints specified in the prompt, avoids contextual contradictions or abrupt logical transitions, and remains consistent with common knowledge for non-creative tasks. The score is defined as

\begin{equation}
\mathrm{TL}=
\begin{cases}
1, & T^m \text{ is judged coherent},\\
0, & \text{otherwise}.
\end{cases}
\label{eq:TL}
\end{equation}

\paragraph{Image Aesthetic (IA).}
Image Aesthetic measures the perceptual quality of generated images using RALI~\citep{rali}. Given a raw aesthetic score $a_i\in[1,5]$ for each generated image, we first normalize it to $[0,1]$ and then average over all generated images:

\begin{equation}
\mathrm{IA}=
\frac{1}{N}
\sum_{i=1}^{N}
\frac{a_i-1}{4}.
\label{eq:IA}
\end{equation}

\paragraph{Image Diversity (ID).}
Image Diversity measures visual repetition between consecutive generated images. We extract normalized DINOv3~\citep{dinov3} features and compute the cosine similarity $s_i$ between each pair of adjacent images. Rather than rewarding arbitrary visual differences, we only penalize excessive similarity beyond a threshold $\tau=0.85$. The score is determined by the most repetitive adjacent pair:

\begin{equation}
\mathrm{ID}
=
\begin{cases}
1, & s_{\max}\leq\tau,\\[2mm]
\dfrac{1-s_{\max}}{1-\tau}, & s_{\max}>\tau,
\end{cases}
\qquad
s_{\max}=\max_i
\operatorname{cos}
\left(
f(I_i^{m}),f(I_{i+1}^{m})
\right).
\label{eq:ID}
\end{equation}

\paragraph{Quantity Compliance (QC).}
Quantity Compliance evaluates whether the model generates the expected number of images. Let $N_m$ and $N_g$ denote the numbers of generated and reference images, respectively. We adopt strict exact matching:

\begin{equation}
\mathrm{QC}
=
\begin{cases}
1, & N_m=N_g,\\
0, & N_m\neq N_g.
\end{cases}
\label{eq:QC}
\end{equation}

\paragraph{Context Fidelity (CF).}
Context Fidelity measures whether a generated image satisfies the visual requirements implied by its preceding interleaved context. For each position $i$, we reuse the context-derived visual rubrics constructed during evaluation-set curation and ask a VLM to verify them against the generated image. Let $R_i$ denote the number of rubrics and $K_i$ the number of satisfied rubrics. We define the position-level score as:

\begin{equation}
\mathrm{CF}_i=
\begin{cases}
0, & K_i=0,\\[1mm]
0.5+\dfrac{K_i-1}{R_i-1}\times0.5, & K_i\geq1.
\end{cases}
\label{eq:CF_i}
\end{equation}

The sample-level score is then

\begin{equation}
\mathrm{CF}
=
\frac{1}{N}
\sum_{i=1}^{N}\mathrm{CF}_i.
\label{eq:CF}
\end{equation}

This scoring function explicitly distinguishes complete contextual failure from partial satisfaction: satisfying none of the expected visual properties receives zero, while satisfying at least one establishes partial contextual grounding, with the score increasing toward one as additional rubrics are fulfilled. In practice, $R_i\geq 2$ for all positions in the evaluation set, so~\cref{eq:CF_i} is always well-defined.

\paragraph{Subject Consistency (SC).}
Subject Consistency evaluates whether the subject generated at each position resembles the corresponding subject in the real-world next image. This metric is evaluated under the fine-grained next-image generation setting, where the model and the real-world sample share the same preceding interleaved context. We therefore use the position-aligned real image as the visual reference and compute the cosine similarity between DINOv3~\citep{dinov3} features of the generated and reference images:

\begin{equation}
\mathrm{SC}
=
\frac{1}{N}
\sum_i
\operatorname{cos}
\left(
f(I_i^{m}),
f(I_i^{g})
\right).
\label{eq:SC}
\end{equation}

A higher score indicates that the generated subject better preserves the visual identity or appearance of the subject that naturally occurs at the corresponding position in the real-world sequence. We apply this metric to Story and Tutorial, where persistent or semantically corresponding visual subjects are meaningful.

\paragraph{Overall Score.}
For each metric, we first average scores over all samples for which the metric is valid. The overall IntReal score is the equally weighted mean of the seven metric scores:

\begin{equation}
\mathrm{Overall}
=
\frac{
\mathrm{TR}+\mathrm{TL}+\mathrm{IA}+\mathrm{ID}
+\mathrm{QC}+\mathrm{CF}+\mathrm{SC}
}{7}.
\label{eq:intreal}
\end{equation}

We emphasize that this overall score is intended primarily as a convenient reference for coarse-grained comparison across models, rather than as a definitive measure of generation quality. Since interleaved generation is a highly multi-faced and open-ended task, we recommend examining performance along individual dimensions to better understand a model's specific strengths and weaknesses, rather than relying solely on the aggregated score.

\subsection{Evaluation Prompts}
\label{app:evaluation-prompts}

\begin{promptbox}{Prompt for Text Richness Evaluation}
You are an expert evaluator comparing two text responses to the same prompt.

\textbf{Task:}
Given a PROMPT and two RESPONSE texts, evaluate which response is better in terms of informativeness, completeness, and conciseness.

\textbf{Evaluation Criteria:}

1. \textit{Informativeness:} Does the response provide meaningful and useful content that genuinely helps the user?

2. \textit{Completeness:} Does the response adequately address the full scope of the prompt without leaving key aspects unaddressed?

3. \textit{Conciseness:} Does the response avoid unnecessary repetition, filler content, or verbosity that dilutes the core information?

\textbf{Important Rules:}

- Evaluate ONLY the textual components. Ignore images or image placeholders.

- Compare the two responses directly against each other.

- Do NOT let response length bias your judgment. A shorter response can be better if it is more precise.

- You MUST pick a winner (1 or 2) unless the two responses are truly indistinguishable in quality. Only output winner: 0 as a last resort when no meaningful difference exists.

\textbf{Output Format:}

\{"winner": 1\} if RESPONSE1 is better.

\{"winner": 2\} if RESPONSE2 is better.

\{"winner": 0\} if the two are roughly equal in quality.

\end{promptbox}

\begin{promptbox}{Prompt for Text Logicality Evaluation}
You are an evaluator for contextual consistency in interleaved text-image generation tasks.

\textbf{Task:}
Determine whether the MODEL RESPONSE is logically consistent, coherent, and free from hallucinations or logical leaps based on the PROMPT and contextual common sense.

\textbf{Evaluation Criteria:}

1. \textit{Context consistency:} The response must strictly follow the context, entities, facts, and constraints provided in the prompt. It should not contradict prior information.

2. \textit{Logical coherence:} The response must be a smooth and sensible continuation. It should focus on the same topic/event without abrupt topic shifts or sudden logical leaps.

3. \textit{Real-world Factuality:} For creative/fictional tasks (e.g., storytelling, sci-fi), allow for imaginative elements, judging only by internal consistency. For non-creative/factual tasks, the response must not severely contradict common sense or real-world logic.

\textbf{Important:} The model response is extracted from an interleaved text-image sequence. It may contain placeholders like `[image]` or other special tokens. You should IGNORE these placeholders and evaluate ONLY the logical coherence and factuality of the actual text content.

\textbf{Output Format:}
You must output ONLY a single valid JSON object and nothing else. No explanations, no analysis, no markdown.

\{"coherent": true\} if the model response is consistent.

\{"coherent": false\} if the model response is inconsistent.
\end{promptbox}

\begin{promptbox}{Prompt for Context Fidelity Evaluation}
You are an image checker. Given an image and a numbered checklist of rubric items, verify each item against what you actually see in the image.

\textbf{For each rubric item:}

- Mark true if the described object, action, state, or property is satisfied.

- Mark false if the item is clearly absent or completely contradicted by the image.

- Err on the side of false when in doubt.

\textbf{Output EXACTLY:}
\{"passes": [true, false, true, ...]\} The boolean array MUST have the same length as the number of rubric items (one boolean per item, in order).

\textbf{CRITICAL RULES:}

- Output ONLY the JSON object. No markdown, no code fences, no preamble, no extra text.

- The "passes" array must contain only booleans (true/false, not strings).

- Array length MUST exactly equal the number of rubric items.
\end{promptbox}

\subsection{Metric Calibration Cases}
We present representative low-scoring examples from different models to illustrate the failure patterns captured by each metric.

\begin{figure*}[ht!]
  \centering
   \includegraphics[width=\linewidth]{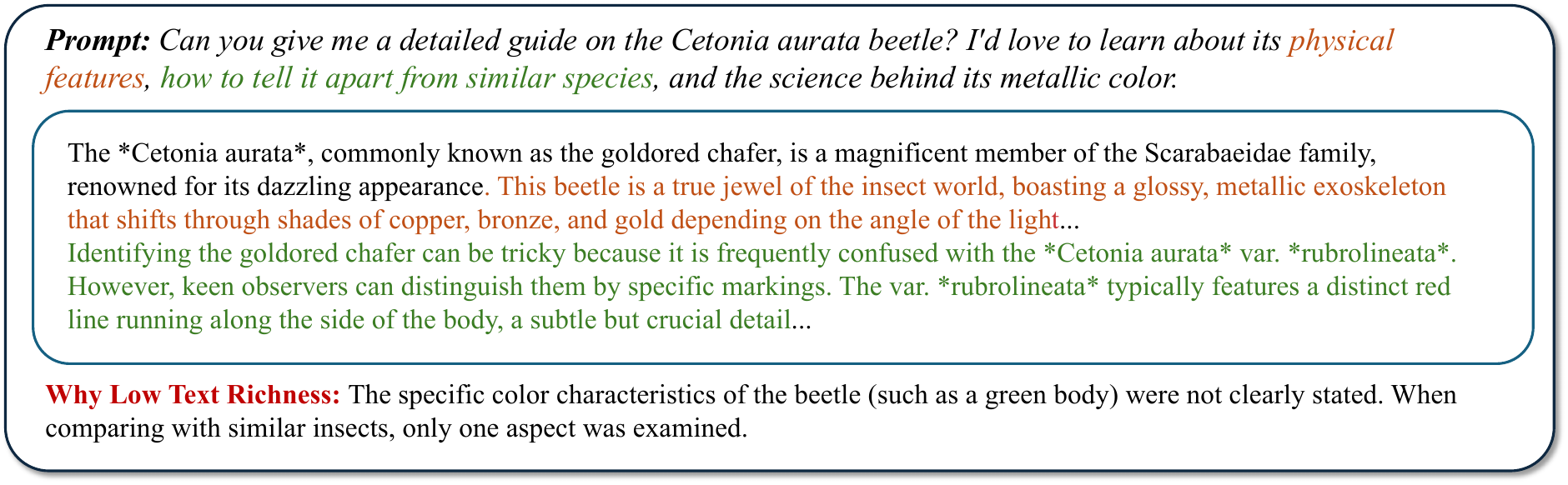}
   \caption{\textbf{Calibration case for Text Richness.}
A low-scoring sample from \textbf{Bagel+SFT}, where the response fails to provide sufficiently specific insect characteristics and offers only limited detail when distinguishing the target insect from similar species, leading to a low \textbf{Text Richness} score.}
   \label{fig:calib-TR}
\end{figure*}

\begin{figure*}[t!]
  \centering
   \includegraphics[width=\linewidth]{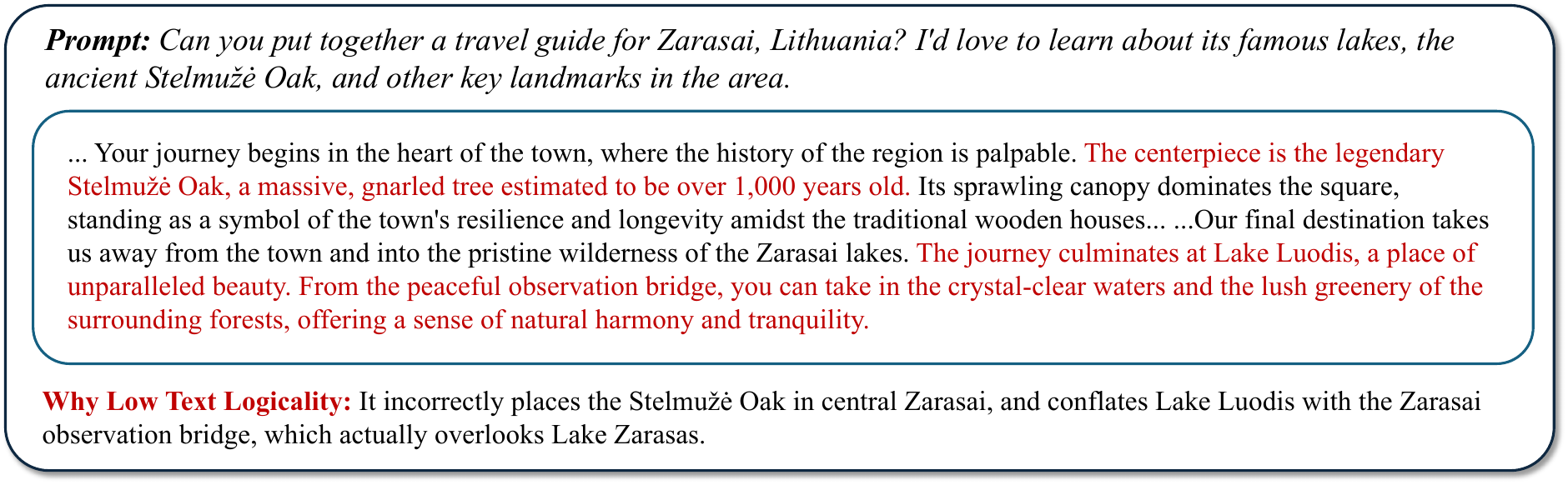}
   \caption{\textbf{Calibration case for Text Logicality.}
A low-scoring sample from \textbf{U1-8B-MoT}, where spatial and factual inconsistencies in the description lead to a low \textbf{Text Logicality} score.}
   \label{fig:calib-TL}
\end{figure*}

\begin{figure*}[t!]
  \centering
   \includegraphics[width=\linewidth]{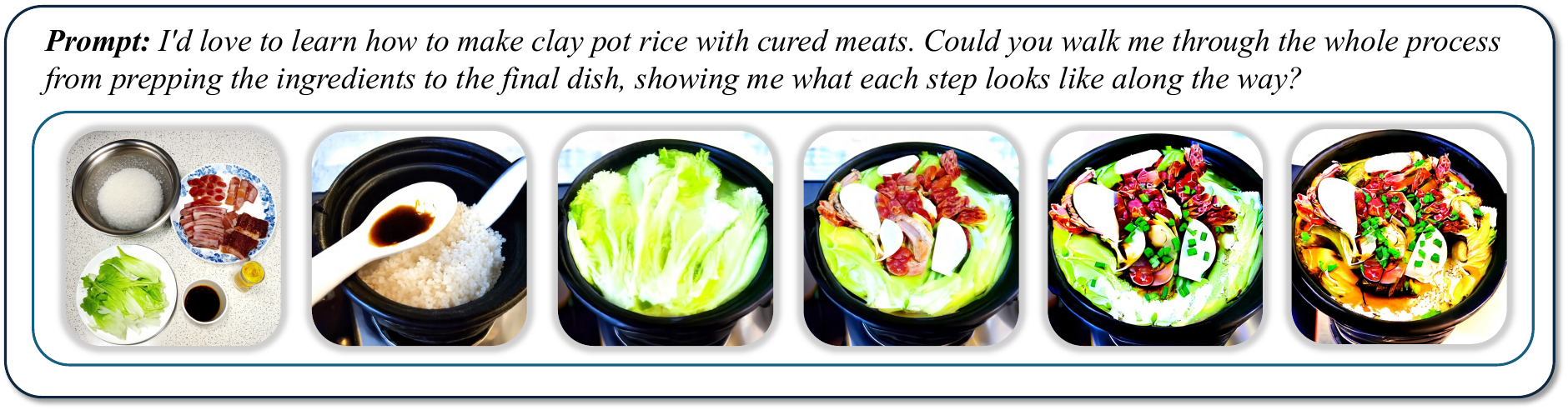}
   \caption{\textbf{Calibration case for Image Aesthetic.}
A low-scoring sample from \textbf{U1-8B-MoT-SFT}, where image quality degrades over steps, with later images exhibiting excessively strong saturation and contrast and appearing less photorealistic, leading to a low \textbf{Image Aesthetic} score.}
   \label{fig:calib-IA}
\end{figure*}

\begin{figure*}[t!]
  \centering
   \includegraphics[width=\linewidth]{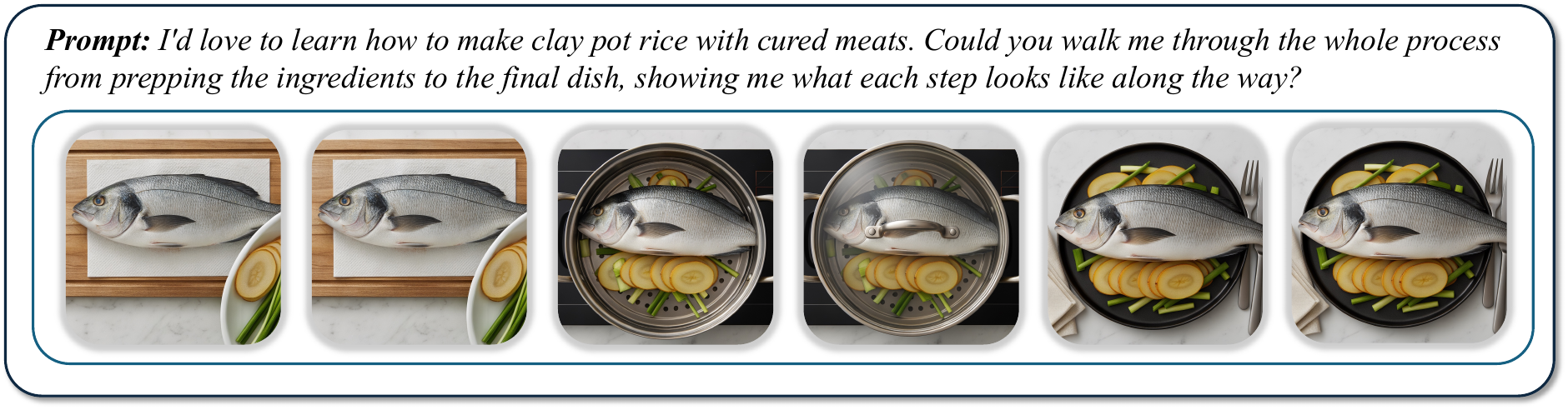}
   \caption{\textbf{Calibration case for Image Diversity.}
A low-scoring sample from \textbf{Emu3.5}, where Images 1\&2, 3\&4, and 5\&6 are excessively similar to each other, resulting in limited visual variation and a low \textbf{Image Diversity} score.}
   \label{fig:calib-ID}
\end{figure*}

\begin{figure*}[t!]
  \centering
   \includegraphics[width=\linewidth]{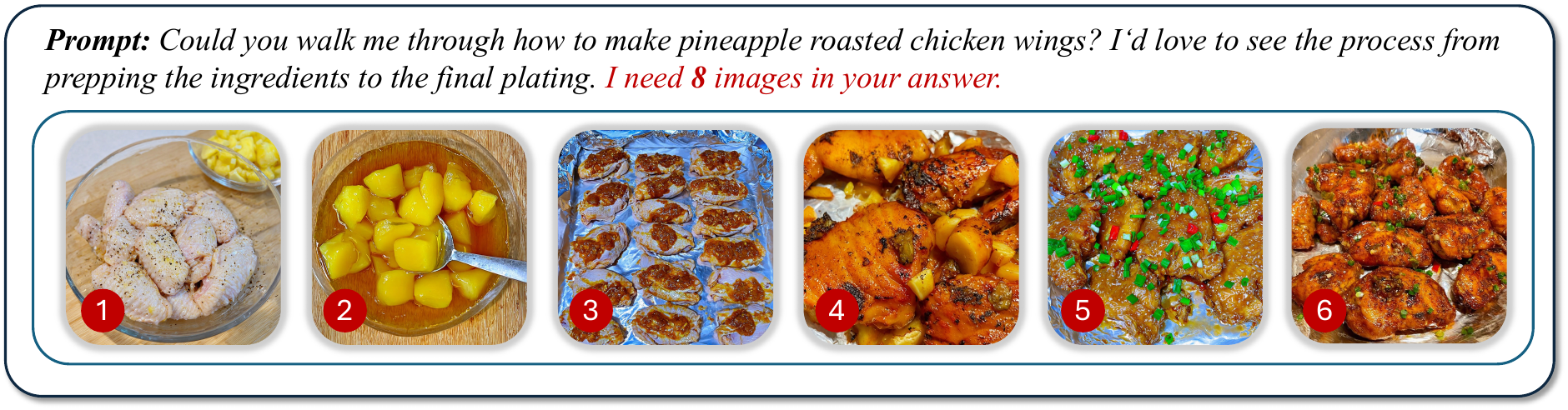}
   \caption{\textbf{Calibration case for Quantity Compliance.}
A low-scoring sample from \textbf{Bagel+SFT+RL}, where the prompt requests eight images but the response generates only six, leading to a low \textbf{Quantity Compliance} score.}
   \label{fig:calib-QC}
\end{figure*}

\begin{figure*}[t!]
  \centering
   \includegraphics[width=\linewidth]{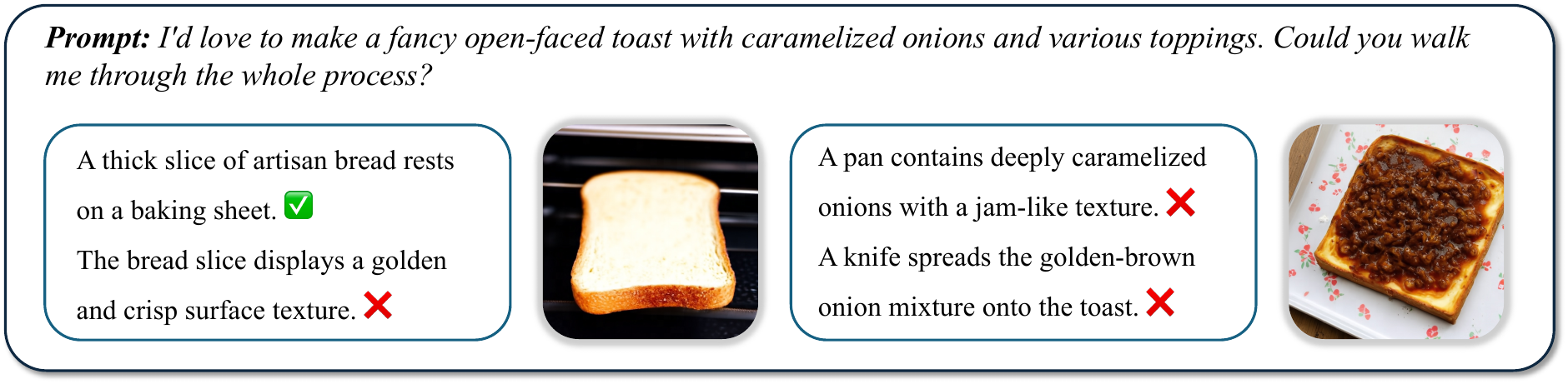}
   \caption{\textbf{Calibration case for Context Fidelity.}
A low-scoring sample from \textbf{U1-8B-MoT+SFT+RL}, where the generated images fail to faithfully follow the described cooking process: the bread is insufficiently toasted in the first image, while the second image omits the preparation and spreading of the caramelized onions, leading to a low \textbf{Context Fidelity} score.}
   \label{fig:calib-CF}
\end{figure*}

\begin{figure*}[t!]
  \centering
   \includegraphics[width=\linewidth]{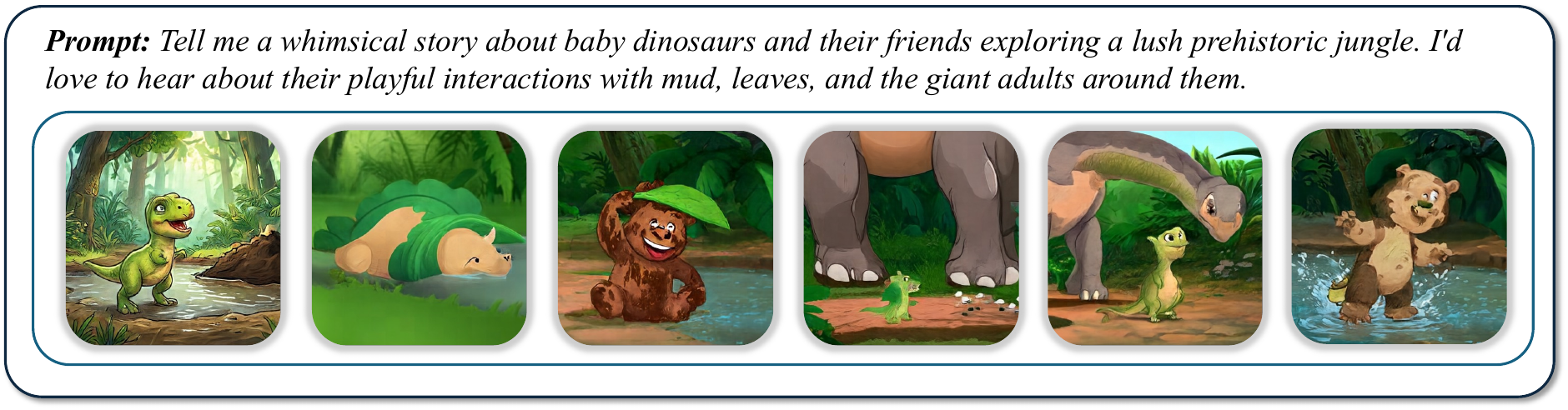}
   \caption{\textbf{Calibration case for Subject Consistency.}
A low-scoring sample from \textbf{U1-A3B-MoT}, where the main subjects are not consistently preserved across images. For example, the green dinosaur differs across Images 1, 4, and 5, while the teddy bear changes color between Images 3 and 6. Since the reference sequence maintains a stable subject appearance throughout, such internal inconsistency in the generated images necessarily manifests as low similarity to the reference at the corresponding positions, leading to a low Subject Consistency score.}
   \label{fig:calib-SC}
\end{figure*}

\begin{figure}[t]
    \centering
    \begin{minipage}[t]{0.48\linewidth}
        \centering
        \adjustbox{valign=t}{\includegraphics[width=\linewidth]{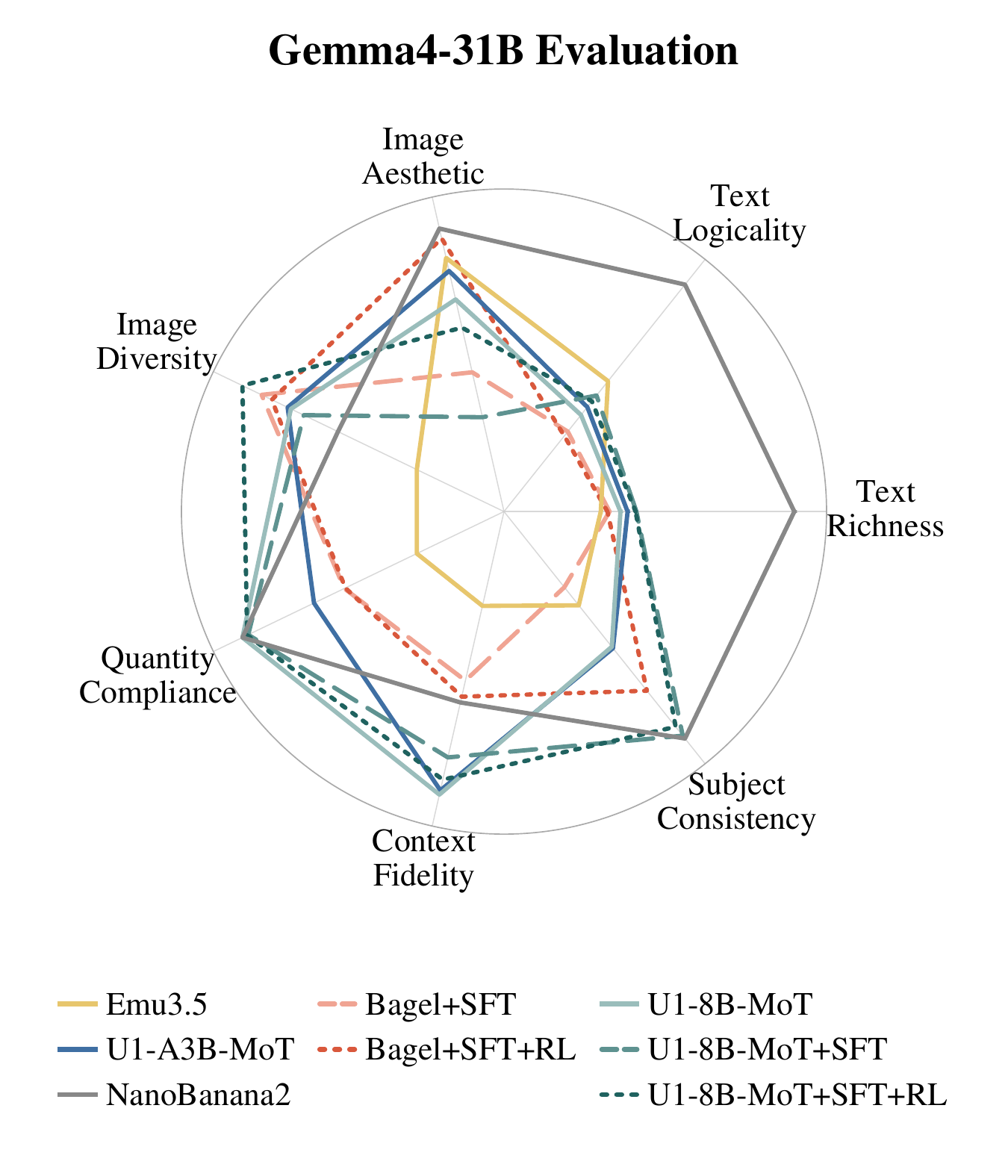}}
        \caption{\textbf{IntReal evaluation results with Gemma4-31B as the evaluator.} We compare models across seven evaluation dimensions.}
        \label{fig:radar-gemma}
    \end{minipage}
    \hfill
    \begin{minipage}[t]{0.48\linewidth}
        \centering
        \adjustbox{valign=t}{\includegraphics[width=\linewidth]{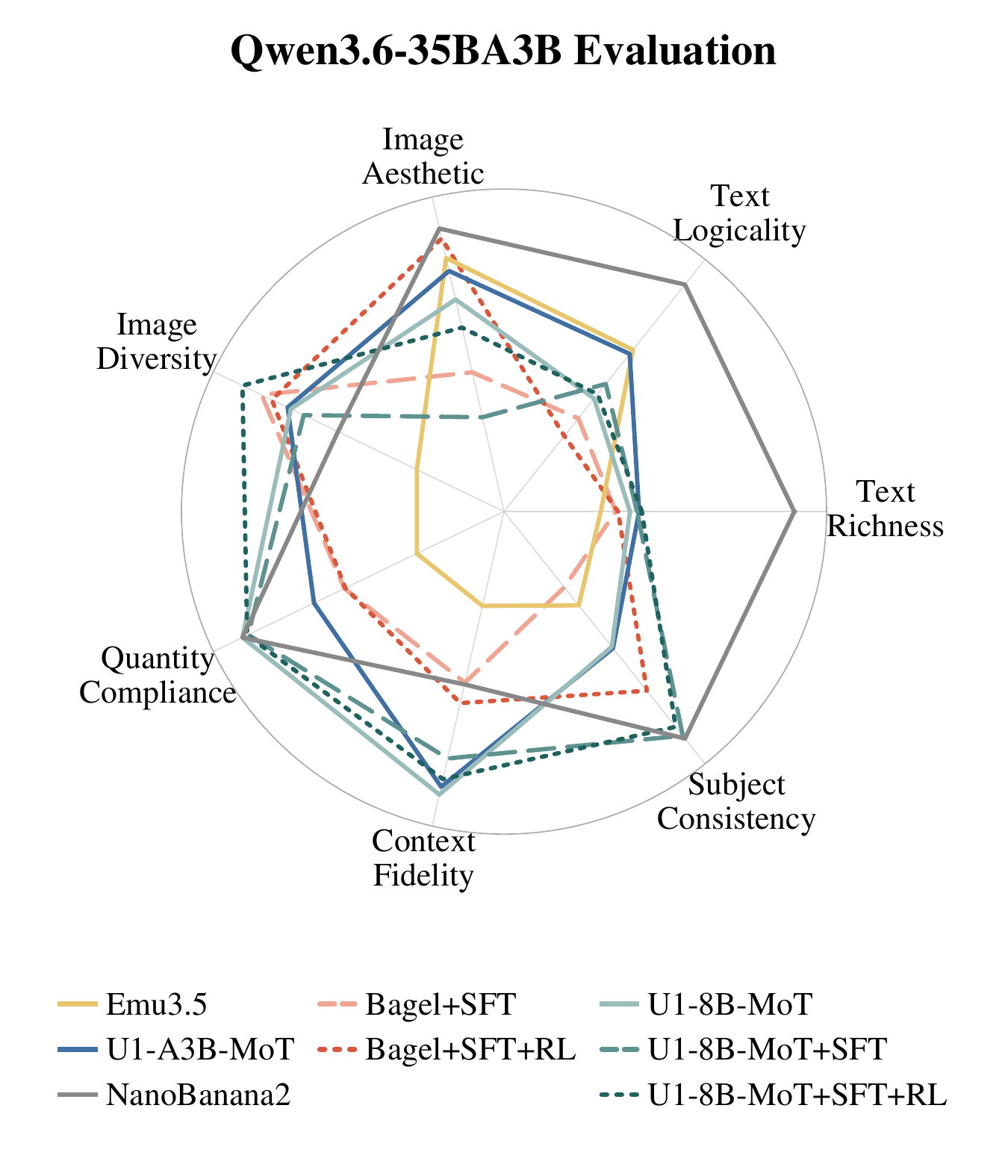}}
        \caption{\textbf{IntReal evaluation results with an alternative evaluator.} We use Qwen3.6-35B-A3B in place of Gemma4-31B as the evaluator.}
        \label{fig:radar-qwen}
    \end{minipage}
\end{figure}

\clearpage
\section{Experiment Details}
\label{sec:app-exp}

\subsection{Model Training Settings}

\begin{table*}[t]
\centering
\caption{\textbf{SFT configurations for Bagel and SenseNova-U1.}
We report the main optimization, batching, loss, and distributed training hyperparameters used in our post-training experiments.}
\begin{tabular}{lcc}
\toprule
\textbf{Hyperparameter} 
& \textbf{Bagel} & \textbf{SenseNova-U1} \\
\midrule
Initialization & BAGEL-7B-MoT & SenseNova-U1-8B-MoT-Interleaved \\
Training steps & 6,700 & 7,000 \\
Optimizer & AdamW & AdamW \\
Learning rate & $2\times10^{-5}$ & $2\times10^{-5}$ \\
LR schedule & Cosine & Cosine \\
Warmup steps & 670 & 700 \\
Adam $\beta_1 / \beta_2$ & 0.9 / 0.95 & 0.9 / 0.95 \\
Adam $\epsilon$ & $1\times10^{-15}$ & $1\times10^{-8}$ \\
Weight decay & 0 & 0 \\
Gradient clipping norm & 1.0 & 1.0 \\
Gradient accumulation steps & 4 & 4 \\
Microbatch size / GPU & 1 packed sequence & 1 packed sequence \\
Packed Sequence Length & 36,000 & 18,000 \\
Text / Image Loss Weights & 1.0 / 1.0 & 1.0 / 1.0 \\
EMA decay & 0.995 & 0.999 \\
Distributed configuration & FSDP, 8 shards & ISP, WP=8, TP=1, PP=1, ZeRO-1 \\
\bottomrule
\end{tabular}
\label{tab:sft-config}
\end{table*}

\begin{table*}[t]
\centering
\caption{\textbf{RL configurations for Bagel and SenseNova-U1.}
We report the main rollout, stochastic sampling, reward, and evaluation configurations used in our reinforcement learning experiments.}
\begin{tabular}{lcc}
\toprule
\textbf{Hyperparameter} & \textbf{Bagel} & \textbf{SenseNova-U1} \\
\midrule
Training resolution & 1024 & 1024 \\
Rollout denoising steps & 10 & 20 \\
Stochastic scheduler & Flow-CPS & Flow-SDE \\
Noise level & 0.7 & 0.4 \\
Candidate stochastic step indices & $\{1,2,3\}$ & $\{1,2,3\}$ \\
Selected stochastic steps / rollout & 1 & 1 \\
Rollout text CFG & 1.0 & 4.0 \\
Rollout image CFG & 1.0 & 1.0 \\
Reward weights (RALI / DINO / Rubric) & 1.5 / 0.5 / 1.0 & 1.0 / -- / -- \\
Evaluation denoising steps & 50 & 50 \\
\bottomrule
\end{tabular}
\label{tab:rl-config}
\end{table*}

We build on the official codebases of Bagel and SenseNova-U1 for SFT, with their hyperparameters summarized in Table~\ref{tab:sft-config}. The two models use different visual representations and consequently organize their visual tokens in the KV cache differently. In Bagel, reference images in the prompt are represented by both VAE latent tokens and ViT embeddings, whereas generated images use only VAE latent tokens. SenseNova-U1 instead performs visual understanding and generation in pixel space, using representations derived from image patches. To accommodate their different visual token budgets, we adopt model-specific smart resizing strategies that preserve aspect ratios and align spatial dimensions to the required grids. For Bagel, the VAE pathway uses a maximum-edge limit of 768 pixels and a target minimum short edge of 512 pixels, with dimensions aligned to multiples of 16; the ViT pathway uses corresponding settings of 518, 224, and 14. For SenseNova-U1, native-resolution resizing constrains image areas to $512^2$--$1024^2$ pixels for understanding and $768^2$--$1024^2$ pixels for generation, with dimensions aligned to the required patch grid.

Detailed RL hyperparameters are provided in Table~\ref{tab:rl-config}. We initialize Bagel and SenseNova-U1 from their SFT checkpoints at steps 4,500 and 6,900, respectively, using non-EMA weights. Given the limited SFT data and training steps, we found that EMA weights lagged behind the updated language branch, resulting in underfitting. We then perform RL post-training using Flow-Factory. For Bagel, we adopt Flow-CPS~\cite{flowcps} as the stochastic sampler. For SenseNova-U1, we retain Flow-SDE, since Flow-CPS is not implemented for U1 in our training setup and its applicability to pixel-space diffusion remains unverified. We set the stochastic noise level $\alpha$ to 0.7 and 0.4 for Bagel and SenseNova-U1, respectively. Following MixGRPO~\citep{mixgrpo}, we accelerate training by randomly selecting a single step among the first three denoising steps for an SDE transition and policy optimization, while using deterministic transitions elsewhere. For stability, Bagel uses no CFG during rollout or policy updates. For SenseNova-U1, we adopt Hybrid CFG following Qwen-Image-2.0-RL~\citep{qwenimagerl}, which we found more effective in our preliminary experiments. We emphasize aesthetic rewards for both models, motivated by our observation that improved visual quality and more faithful object rendering are accompanied by better text-image alignment and subject consistency. For experiments involving multiple reward signals, we adopt GDPO~\citep{gdpo} to balance and optimize multiple reward objectives.

SFT experiments are conducted on a single node of 8×H800-80G GPUs. RL experiments are conducted on two nodes of 8×H800-80G GPUs (16 GPUs in total) to accelerate policy optimization.

\subsection{Model Inference Settings}

During inference, models generate text autoregressively and switch to image generation when the special \texttt{<image>} token is produced. We follow the reasoning and generation template used by Emu3.5~\citep{emu3p5} for interleaved generation. Unless otherwise specified, all generated images are produced at a resolution of $1024\times1024$ for both holistic and fine-grained evaluation. We recommend $1024\times1024$ as the default inference resolution, as it provides sufficient visual fidelity while avoiding excessive growth in visual tokens and the resulting interleaved context length.

Most models are successfully evaluated on all 600 samples in the evaluation set. However, for Emu3.5, eight samples exceed the maximum context length or memory limits even on 8×H800-80G GPUs, so we report its results on the remaining 592 samples. For NanoBanana2, two samples repeatedly triggered the API's built-in safety guardrails and could not be processed, so we report its results on the remaining 598 samples. Manual inspection revealed no apparent safety concerns in these two samples, suggesting that the guardrails may have been falsely triggered.

\subsection{Detailed Evaluation Results}

\begin{table*}[t!]
\centering
\caption{\textbf{IntReal evaluation results with an alternative evaluator.} We report results following the protocol in~\cref{tab:main-results}, using Qwen3.6-35B-A3B in place of Gemma4 as the evaluator.}
\resizebox{\textwidth}{!}{
\begin{tabular}{l|ccccc|cc|c}
\toprule
\multicolumn{1}{c|}{\multirow{3}{*}{\textbf{Model}}}
& \multicolumn{5}{c|}{\textbf{Holistic Evaluation}}
& \multicolumn{2}{c|}{\textbf{Fine-grained Evaluation}}
& \multirow{3}{*}{\textbf{Mean}} \\
\cmidrule(lr){2-6} \cmidrule(lr){7-8}
& \makecell[c]{Text\\Richness}
& \makecell[c]{Text\\Logicality}
& \makecell[c]{Image\\Aesthetic}
& \makecell[c]{Image\\Diversity}
& \makecell[c]{Quantity\\Compliance}
& \makecell[c]{Context\\Fidelity}
& \makecell[c]{Subject\\Consistency} & \\
\midrule
Emu3.5 & 13.34 & \textbf{69.93} & 81.20 & 75.51 & 39.70 & 71.06 & 61.46 & 58.89 \\
\midrule
U1-A3B-MoT & 29.50 & 69.00 & 80.70 & 88.70 & 72.50 & 89.11 & 63.45 & 70.42 \\
\midrule
Bagel\textit{+SFT} & 19.67 & 54.33 & 76.82 & 91.35 & 62.83 & 78.71 & 60.61 & 63.48 \\
Bagel\textit{+SFT+RL} & 20.67 & 50.33 & \textbf{81.94} & 90.31 & 62.50 & 80.76 & 65.43 & 64.56 \\
\midrule
U1-8B-MoT & 25.67 & 58.83 & 79.60 & 88.36 & \textbf{95.33} & \textbf{89.88} & 63.38 & 71.58 \\
\quad\textit{+SFT} & 28.67 & 62.17 & 75.10 & 87.05 & 93.50 & 86.28 & \textbf{67.52} & 71.47 \\
\(\Delta\) vs. base (\%)
& \cellcolor{DeltaGreen!100.0}\textit{+11.69\%}
& \cellcolor{DeltaGreen!56.8}\textit{+5.68\%}
& \cellcolor{DeltaRed!56.5}\textit{\textminus5.65\%}
& \cellcolor{DeltaRed!14.8}\textit{\textminus1.48\%}
& \cellcolor{DeltaRed!19.2}\textit{\textminus1.92\%}
& \cellcolor{DeltaRed!41.1}\textit{\textminus4.11\%}
& \cellcolor{DeltaGreen!65.3}\textit{+6.53\%}
& \cellcolor{DeltaRed!1.5}\textit{\textminus0.15\%} \\
\quad\textit{+SFT+RL} & \textbf{30.67} & 59.83 & 78.53 & \textbf{93.31} & 93.83 & 88.34 & 67.11 & \textbf{73.09} \\
\(\Delta\) vs. base (\%)
& \cellcolor{DeltaGreen!100.0}\textit{+19.48\%}
& \cellcolor{DeltaGreen!17.0}\textit{+1.70\%}
& \cellcolor{DeltaRed!13.4}\textit{\textminus1.34\%}
& \cellcolor{DeltaGreen!56.0}\textit{+5.60\%}
& \cellcolor{DeltaRed!15.7}\textit{\textminus1.57\%}
& \cellcolor{DeltaRed!18.2}\textit{\textminus1.82\%}
& \cellcolor{DeltaGreen!58.9}\textit{+5.89\%}
& \cellcolor{DeltaGreen!21.1}\textit{+2.11\%} \\
\midrule
\rowcolor{gray!7}
NanoBanana2 & 94.98 & 84.78 & 82.32 & 83.56 & 95.32 & 78.89 & 67.65 & 83.93 \\
\rowcolor{gray!7}
$\Delta$ vs. OSS Best (\%)
& +209.68\%
& +21.24\%
& +0.46\%
& -10.45\%
& -0.01\%
& -12.32\%
& +0.19\%
& +14.83\% \\
\bottomrule
\end{tabular}
}
\label{tab:main-results-qwen}
\end{table*}

\paragraph{Evaluation results with Qwen3.6.} We repeat the evaluation in~\Cref{tab:main-results}, replacing Gemma4 with Qwen3.6-35B-A3B as the evaluator. The resulting scores are reported in~\Cref{tab:main-results-qwen}. Across most dimensions, the two evaluators produce nearly identical scores, and model rankings remain highly consistent, corroborating the robustness analysis in~\cref{sec:exp-reliability}. The one notable exception is Text Logicality, which shows larger absolute discrepancies between the two evaluators (e.g., a 12.71-point gap for NanoBanana2). We attribute this to the binary nature of the Text Logicality judgment: unlike Text Richness, which relies on a direct pairwise comparison against a reference response, Text Logicality requires each evaluator to independently determine whether a response is coherent, without an explicit comparative anchor. As a result, each evaluator's own internal calibration of coherence introduces a systematic offset in absolute scores. Despite this discrepancy in absolute values, the relative ranking of models under Text Logicality remains largely preserved across the two evaluators, indicating that our conclusions are not sensitive to the specific evaluator model used. The radar plots in~\cref{fig:radar-gemma,fig:radar-qwen} further show highly similar performance profiles across the two evaluators, providing visual evidence of the robustness of our evaluation.

\begin{figure*}[t!]
  \centering
   \includegraphics[width=\linewidth]{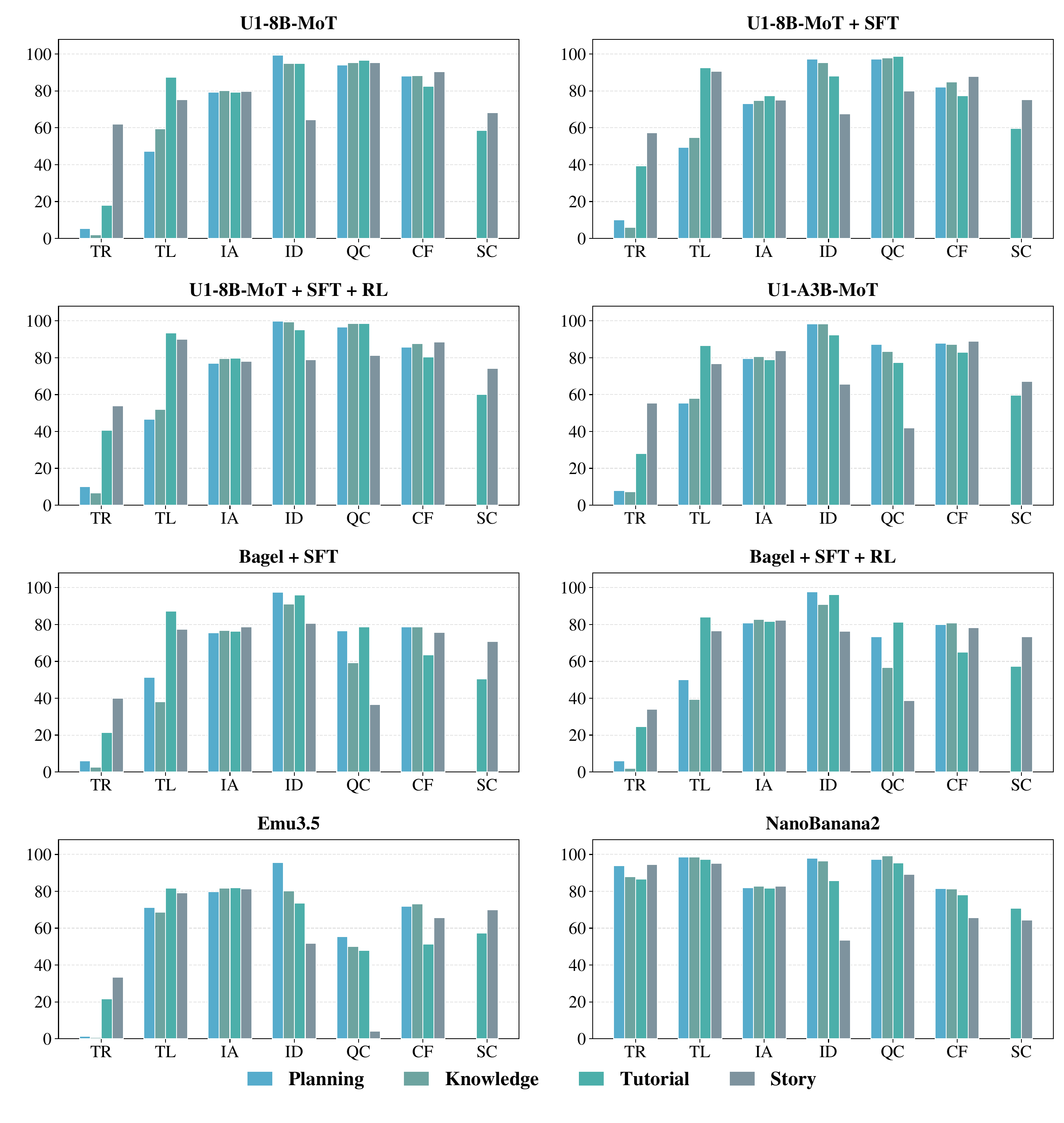}
   \caption{\textbf{IntReal evaluation results per category for individual models.} We report the score of each metric for each category separately for each model.}
   \label{fig:per-cat-per-model}
\end{figure*}

\paragraph{Category-wise Evaluation Results.}
Figure~\ref{fig:per-cat} summarizes the category-wise results of 5 representative models (U1-8B-MoT, Bagel+SFT, U1-A3B-MoT, Emu3.5, and NanoBanana2), while Figure~\ref{fig:per-cat-per-model} provides a detailed breakdown for each model. Open-source models generally exhibit category-wise patterns consistent with the aggregate results: Planning and Knowledge pose greater challenges in Text Richness and Text Logicality, Tutorial is more challenging in Context Fidelity and Subject Consistency, and Story presents greater difficulties in Quantity Compliance and Image Diversity. The closed-source NanoBanana2 exhibits different patterns on some dimensions, particularly Text Richness and Text Logicality, where its consistently strong performance across categories suggests reduced sensitivity to these category-specific challenges.

\section{Qualitative Examples}
\label{sec:app-quant}
\begin{figure*}[ht!]
  \centering
   \includegraphics[width=\linewidth]{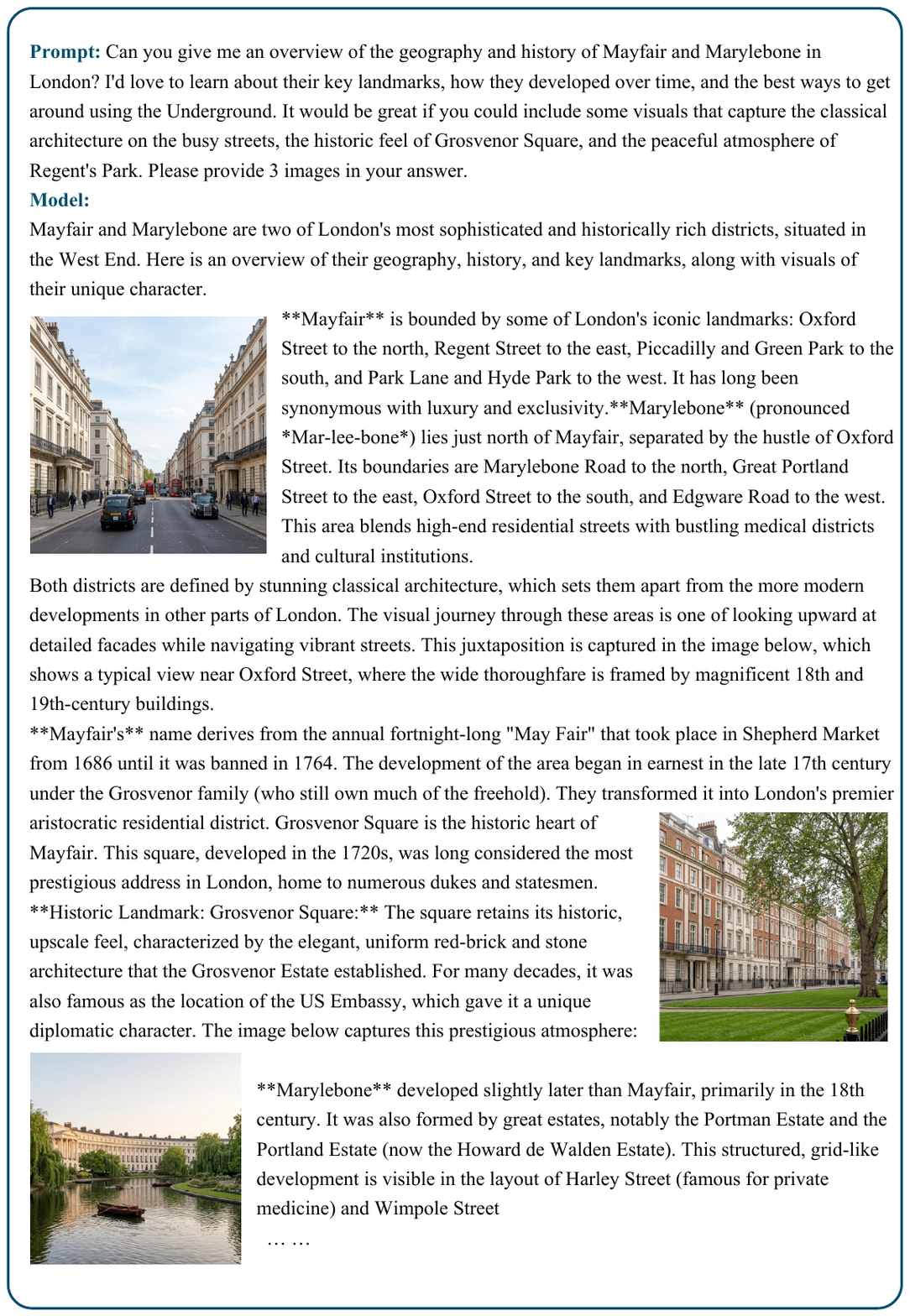}
   \caption{\textbf{Qualitative example from the Planning category generated by Nanobanana2.} The response interleaves text and images to introduce the geography, history, and landmarks of Mayfair and Marylebone.}
   \label{fig:quali-planning}
\end{figure*}

\begin{figure*}[ht!]
  \centering
   \includegraphics[width=\linewidth]{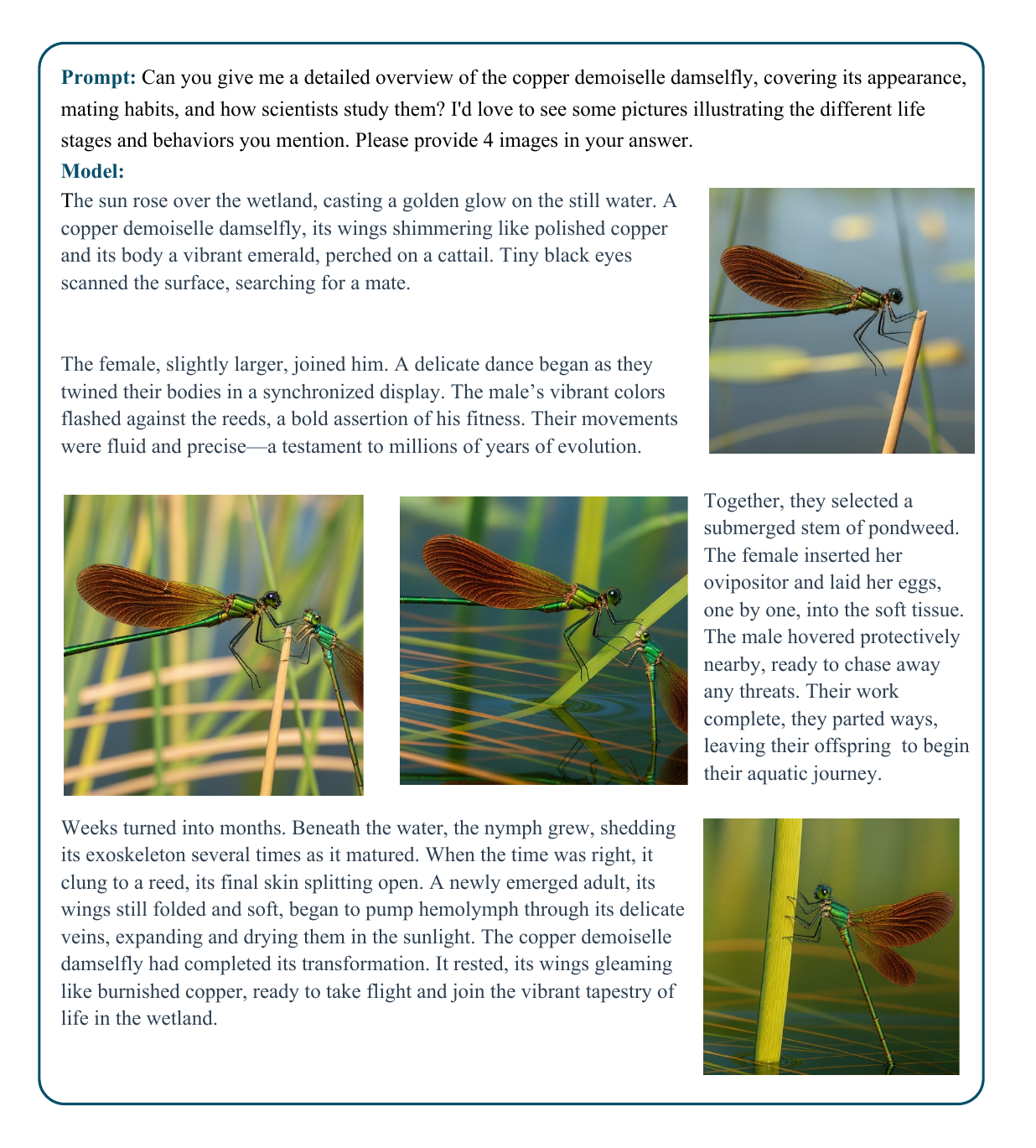}
   \caption{\textbf{Qualitative example from the Knowledge category generated by Emu3.5.} The response interleaves text and images to describe the appearance, reproductive behavior, and development of the copper demoiselle damselfly.}
   \label{fig:quali-knowledge}
\end{figure*}

\begin{figure*}[ht!]
  \centering
   \includegraphics[width=\linewidth]{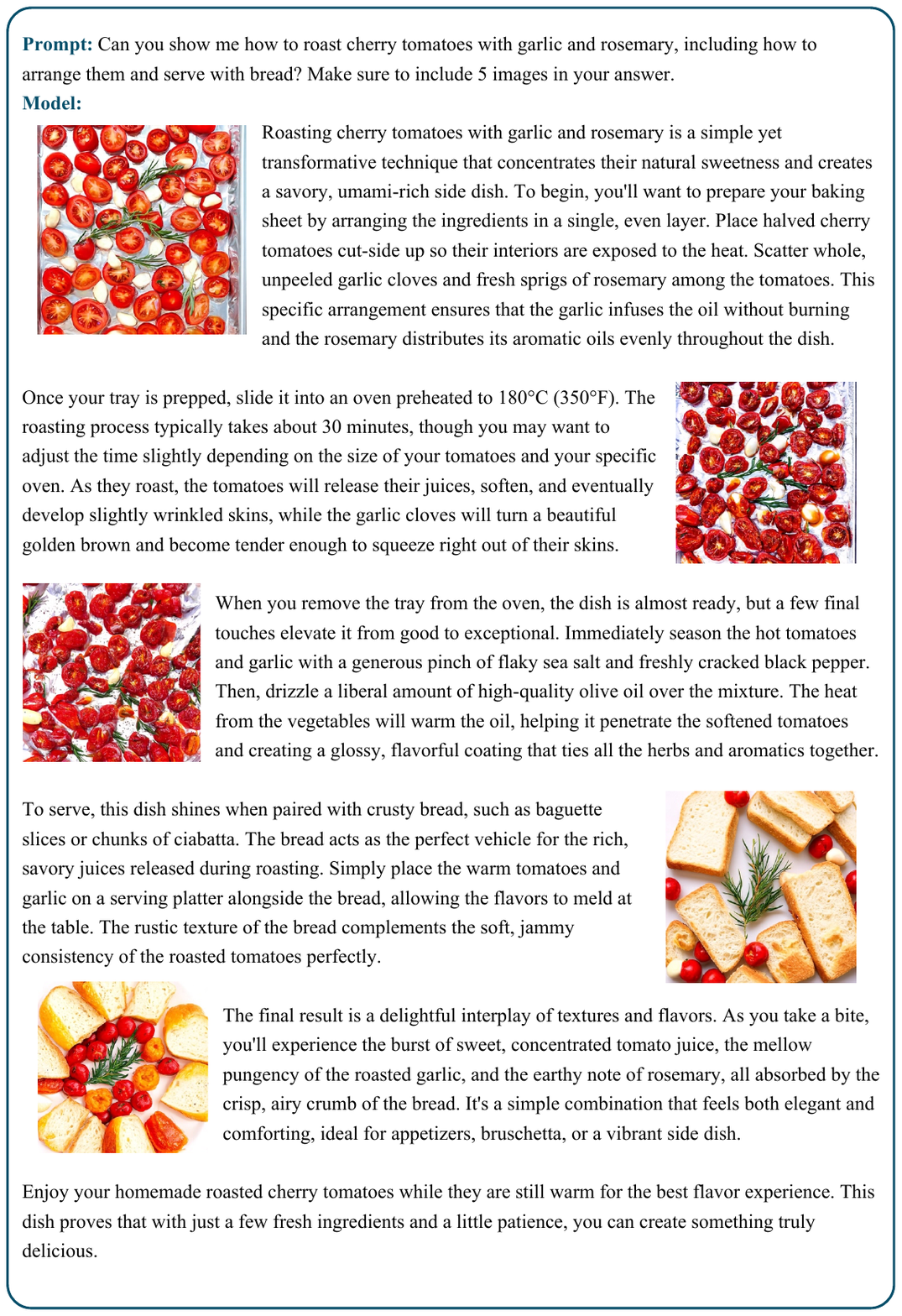}
   \caption{\textbf{Qualitative example from the Tutorial category generated by U1-8B-MoT+SFT+RL.} The response interleaves text and images to explain how to roast cherry tomatoes with garlic and rosemary and serve them with bread.}
   \label{fig:quali-tutorial}
\end{figure*}

\begin{figure*}[ht!]
  \centering
   \includegraphics[width=\linewidth]{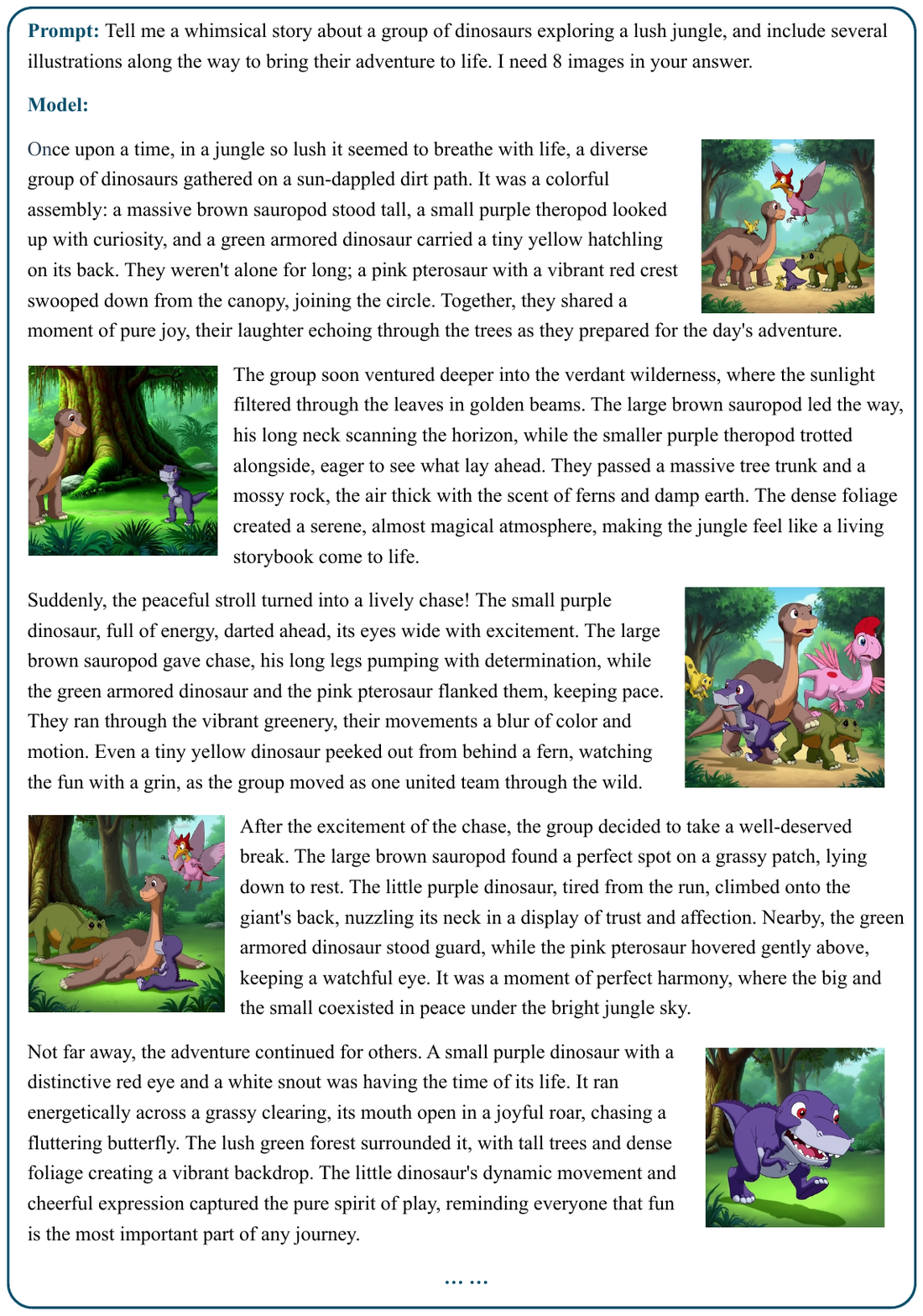}
   \caption{\textbf{Qualitative example from the Story category generated by Bagel+SFT+RL.} The response interleaves text and images to narrate a group of dinosaurs exploring and playing together in a lush jungle.}
   \label{fig:quali-story}
\end{figure*}

\stopcontents[appendix]

\end{document}